\documentclass[10pt, a4paper, twocolumn, teaser, showabstract]{naverlabseurope}

\usepackage{multirow}
\usepackage{multicol}
\usepackage{rotating}
\usepackage{bbm}
\usepackage{algorithm2e,algpseudocode}
\usepackage{hyperref}
\usepackage[capitalise]{cleveref}
\usepackage{url}

\newlength{\smallimage}
        \definecolor{rel}{rgb}{.1,.6,.2}
        \definecolor{nrl}{rgb}{1,1,1}
        \definecolor{qim}{rgb}{1,1,1}

\def\eg{\emph{e.g.}}

\def\ie{\emph{i.e.}}

\def\cf{\emph{c.f.\,}}

\def\wrt{w.r.t.\,}

\definecolor{lightgray}{gray}{0.93}

\def\be{\begin{equation}}
\def\ee{\end{equation}}
\def\bea{\begin{eqnarray}}
\def\eea{\end{eqnarray}}
\def\ben{\begin{eqnarray*}}
\def\een{\end{eqnarray*}}

\def\bi{\begin{itemize}}
\def\ei{\end{itemize}}

\newcommand{\btab}[1]{\begin{tabular}{#1}}
\newcommand{\etab}{\end{tabular}}
\newcommand{\ba}[1]{\begin{array}{#1}}
\newcommand{\ea}{\end{array}}

\def\Re{{\rm I\!R}}                            
\def\<{\langle}
\def\>{\rangle}

\newcommand{\bff}{{\bf f}}
\newcommand{\bfg}{{\bf g}}

\newcommand{\bfl}{{\bf l}}

\newcommand{\bfp}{{\bf p}}

\newcommand{\bfu}{{\bf u}}

\newcommand{\bfF}{{\bf F}}

\newcommand{\bfS}{{\bf S}}

\newcommand{\calG}{{\mathcal G}}

\newcommand{\bbone}{{\mathbbm{1}}}

\newcommand{\myparagraph}[1]{\vspace{0.1cm}\noindent\textbf{#1.}}

\newcommand{\PAR}[1]{\vspace{0.1cm}\noindent\textbf{#1.}}



\definecolor{DarkGreen}{rgb}{0.5, 0.9, 0.5}

\newcommand{\supmat}{{Appendix}}

\newcommand{\gsfloc}{{GSFFs-PR}}
\newcommand{\gsf}{{GSFFs}}

\newcommand{\bd}[1]{{#1}}

\title{Gaussian Splatting Feature Fields for Privacy-Preserving Visual Localization}

\authors{Maxime Pietrantoni$^{1,2,3}$ and 
\authsep 
Gabriela Csurka$^3$ and
\authsep 
Torsten Sattler$^2$}

\affiliations{
$^1$ Faculty of Electrical Engineering, Czech Technical University in Prague, \url{\{firstname.lastname\}@cvut.cz}\\
$^2$ Czech Institute of Informatics, Robotics and Cybernetics, Czech Technical University in Prague \\
$^3$ NAVER LABS Europe, \url{\{firstname.lastname\}@naverlabs.com} \\ } 
\contributions{}
\websiteref{}
\website{}
\date{}
 
\teaserfig{\includegraphics[width=0.8\textwidth]{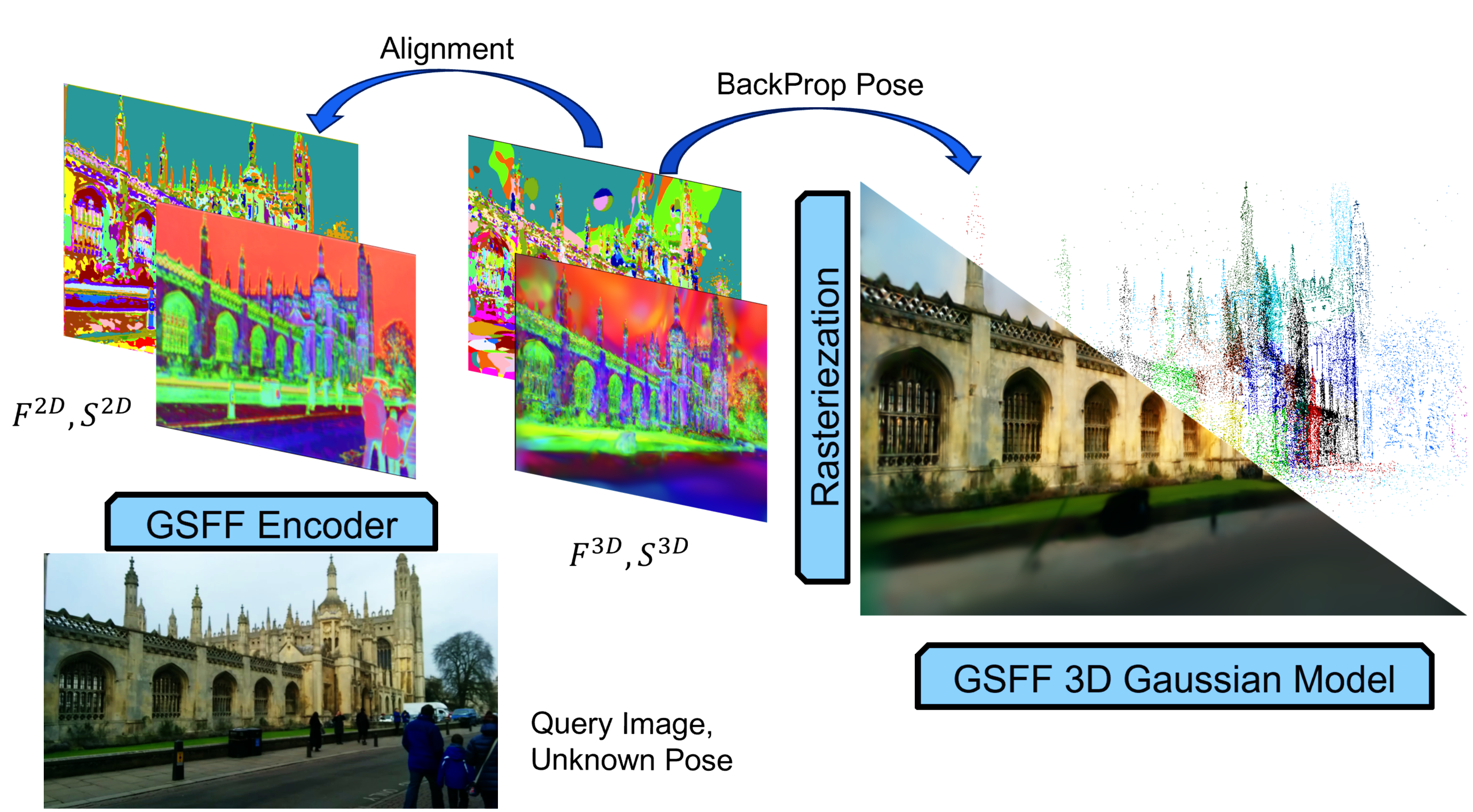}}
\teasercaption{We extract 2D feature $\bfF^\text{2D}$ or segmentation $\bfS^\text{2D}$ maps from a query image with unknown pose. We estimate the pose by aligning the maps with  feature $\bfF^\text{3D}$ or segmentation $\bfS^\text{3D}$ maps rendered from our Gaussian Splatting Feature Fields (GSFFs), which are learned in a self-supervised way. Using segmentations instead of features makes our localization pipeline privacy-preserving.   \label{fig:teaser}}
\date{}

\begin{abstract}
Visual localization is the task of estimating a camera pose in a known environment.  
In this paper, we utilize
3D Gaussian Splatting (3DGS)-based representations for 
accurate and privacy-preserving visual localization. 
We propose Gaussian Splatting Feature Fields (\gsf), a scene representation for visual localization that combines an explicit geometry model (3DGS) with an implicit feature field. 
We leverage the dense geometric information and differentiable rasterization algorithm from 3DGS to learn robust feature representations grounded in 3D. 
In particular, we align a 3D scale-aware feature field and a 2D feature encoder in a common embedding space through a contrastive framework.
Using a 3D structure-informed clustering procedure, we further regularize the representation learning and seamlessly convert the features to segmentations, which can be used for privacy-preserving visual localization. 
Pose refinement, which involves aligning either feature maps or segmentations from a query image with those rendered from the \gsf{} scene representation, is used to achieve localization.
The resulting privacy- and non-privacy-preserving localization pipelines, evaluated on multiple real-world datasets, show state-of-the-art  performances.
\end{abstract}

\begin{document}
\maketitle

\section{Introduction}

Visual localization (VL), a core part of self-driving cars~\cite{HengICRA19ProjectAutoVisionLocalization3DAutonomousVehicle} and autonomous robots~\cite{LimIJRR15RealTimeMonocularImageBased6DoFLocalization}, is the task of estimating the 6DoF camera pose from which an image was captured.

VL methods can be distinguished by how they represent scenes and how they estimate the pose of a query image \wrt  the scene representation. 
Popular choices for the former include  
 3D Structure-from-Motion (SfM) point clouds ~\cite{humenberger2020robuste,SarlinCVPR19FromCoarsetoFineHierarchicalLocalization,
Germain3DV19SparseToDenseHypercolumnMatchingVisLoc,HumenbergerIJCV22InvestigatingRoleImageRetrieval,SarlinCVPR21BackToTheFeature}, databases of images with known intrinsic and extrinsic parameters~\cite{ZhouICRA20ToLearnLocalizationFromEssentialMatrices,Bhayani_2021_ICCV,Zhang06TDPVT,SattlerCVPR19UnderstandingLimitationsPoseRegression}, the weights of neural networks~\cite{KendallICCV15PoseNetCameraRelocalization,BrahmbhattCVPR18GeometryAwareLocalization,MoreauCORL22LENSLocalizationEnhancedByNeRFSynthesis,brachmann2021visual,brachmann2023accelerated}, or dense renderable representations such as meshes~\cite{Panek2022ECCV,trivigno2024unreasonable}, neural radiance fields \cite{MoreauX23CROSSFIRECameraRelocImplicitRepresentation,chen2023refinement,pietrantoni2024jointnerf,zhou2024nerfect}, or 3D Gaussian Splatting~\cite{liu2024gsloc,sidorov2024gsplatloc,zhai2024splatloc,bortolon20246dgs}. 
Common VL approaches 
are based on establishing 2D-3D correspondences via feature matching~\cite{SarlinCVPR19FromCoarsetoFineHierarchicalLocalization,TairaPAMI21InLocIndoorVisualLocalization,Panek2022ECCV,MoreauX23CROSSFIRECameraRelocImplicitRepresentation,zhou2024nerfect} or scene coordinate regression~\cite{brachmann2021visual,brachmann2023accelerated}, relative pose estimation from feature matches~\cite{ZhouICRA20ToLearnLocalizationFromEssentialMatrices,Zheng2015ICCV,Bhayani_2021_ICCV,Zhang06TDPVT} or 
regression~\cite{Ng20223DV,ZhouICRA20ToLearnLocalizationFromEssentialMatrices}, absolute pose regressed via neural networks~\cite{WangAAAI20AtLocAttentionGuidedCameraLocalization,Shuai3DV21DirectPoseNetwithPhotometricConsistency,ShuaiECCV22DFNetEnhanceAPRDirectFeatureMatching,MoreauCORL22LENSLocalizationEnhancedByNeRFSynthesis}, or feature-based pose refinement~\cite{trivigno2024unreasonable,SarlinCVPR21BackToTheFeature,pietrantoni2023segloc} of an initial pose estimate. 
While local feature matching-based approaches provide the most accurate camera pose estimates,
scale to large scenes~\cite{SattlerICCV15HyperpointsFineVocabulariesLocRecogn,HumenbergerIJCV22InvestigatingRoleImageRetrieval}, handle changing conditions~\cite{toft2020long} and fit onto mobile devices~\cite{LynenRSSC15GetOutVisualInertialLocalization,LimIJRR15RealTimeMonocularImageBased6DoFLocalization}, 
they suffer from potential privacy-related issues as image details can be recovered from the feature descriptors~\cite{pittaluga2019revealing,chelani2024obfuscation}.

Since VL 
systems are often deployed through cloud-based solutions, preserving the privacy of user-uploaded images and scenes is a critical aspect~\cite{SpecialeCVPR19PrivacyPreservingImageBasedLocalization,SpecialeICCV19PrivacyPreservingImageQueriesforCameraLocalization}.
An interesting approach to both query and scene privacy is SegLoc~\cite{pietrantoni2023segloc}, which is a pose refinement-based method that represents the scene as a sparse SfM point cloud. 
Most refinement-based methods project features associated with the scene geometry into the query image~\cite{SarlinCVPR21BackToTheFeature,von2020gn,vonstumberg2020lmreloc,pietrantoni2024jointnerf,MoreauX23CROSSFIRECameraRelocImplicitRepresentation,trivigno2024unreasonable}. 
An initial pose estimate, \eg, obtained via image retrieval, is then optimized by aligning the projected features with features extracted from the query image. 
To increase privacy while decreasing 
memory requirements 
for storing high-dimensional features, 
SegLoc ~\cite{pietrantoni2023segloc}  propose to quantize the features into integer values,
which is  equivalent to assigning segmentation labels to pixels in the query image and the 3D scene points. These 
quantized representations leads to better privacy-preservation:  only coarse image information without any details can be recovered from both 2D segmentation images and 3D point clouds with associated labels.
Final poses are refined by maximizing label consistency between the projected points and pixel  labels.

The segmentations used by SegLoc are  learned purely in 2D, providing 
no guarantee that the predicted labels are consistent between viewpoints. 
In contrast, \cite{pietrantoni2024jointnerf}
jointly trains a dense scene representation (in the form of a NeRF) together with an implicit feature field,  ensuring multi-view consistency and 
state-of-the-art feature-based  pose refinement results. 
Yet, \cite{pietrantoni2024jointnerf} 
is not privacy-preserving. 
In contrast, in this work, we  
investigate learning a feature field that can be used for privacy-preserving visual localization jointly with a dense scene representation. 
We chose 3D Gaussian Splatting (3DGS)~\cite{KerblTOG233DGaussianSplatting} due to its fast rendering time, making it particularly suited for dense pose refinement. 
Compared to SegLoc, it allows us to ground representation learning in 3D. 
Additionally, the explicit and finite nature of 3DGS is better suited for privacy-preserving localization than NeRFs 
as, similarly  to~\cite{pietrantoni2023segloc}, we can simply quantize the features associated with the Gaussians. 

Some prior~\cite{bortolon20246dgs} and concurrent works \cite{liu2024gsloc,sidorov2024gsplatloc,zhai2024splatloc} also use 3DGS in the context of visual localization. However, their pipelines use matching-based solutions, whereas we take advantage of the ability to densely render from any viewpoint within the scene to perform feature-metric or segmentation-based pose refinement. We explicitly backpropagate through the rasterizer on the se(3) Lie algebra, which yields more accurate pose estimates. 
Furthermore, instead of relying on pre-trained features as \cite{trivigno2024unreasonable}, we learn our features in a self-supervised manner
defining a 
Gaussian Spatting Feature Field (\gsf{}), which associates 3D Gaussians with volumetric features extracted with a kernel-based encoding based on the covariance of the 3D Gaussians. 
These features are rendered and aligned to  features provided by the 2D encoder (which is jointly trained with the \gsf) through contrastive losses. We apply further regularization by leveraging the geometry of the 3DGS model and by spatially clustering the \gsf. These clusters enables  converting  features into segmentations that can be used to perform effective privacy-preserving pose refinement (\cf Fig.~\ref{fig:teaser}).

To summarize, our contributions are threefold:
1) We introduce Gaussian Feature Fields, a novel representation for VL, jointly 
learnt with the image feature encoder 
enabling  pose refinement by aligning rendered and extracted features.
2) The explicit nature of the 3D representation in \gsf{} allows to spatially cluster the Gaussian cloud and segmenting the feature field based on cluster centers
yielding a privacy-preserving 3D scene representation.
Similar to~\cite{pietrantoni2023segloc}, the resulting method is based on aligning discrete segmentation labels, thus leading to similar privacy-preserving properties.  
3) We demonstrate the accuracy of both 
\gsf-based localization pipelines (based on features respectively segmentations) on multiple real-world datasets. 

Our approaches outperform prior and concurrent work for privacy- and non-privacy-preserving visual localization.

\section{Related Works}

\PAR{Gaussian Splatting}
3D Gaussian Splatting \cite{KerblTOG233DGaussianSplatting} represents a scene as a set of Gaussian primitives and render images by rasterizing Gaussians using depth ordering and alpha-blending, thus allowing efficient training and high resolution real-time rendering. Thanks to these properties, 3D Gaussians have been applied to numerous tasks including SLAM \cite{keetha2024splatam,matsuki2024gaussian}, dynamic scene modeling \cite{luiten2023dynamic,yang2024deformable}, scene segmentation \cite{ji2024segment,liao2024clip}, and surface reconstruction \cite{huang20242d,guedon2024sugar,ZehaoTOG24GaussianOpacityFields}. Extensions have been tackling anti-aliasing \cite{yu2024mip}, sparse view setups \cite{xiong2023sparsegs,yu2024lm}, removing reliance on estimated poses  \cite{ye2024no}, and regularization \cite{chung2024depth,cheng2024gaussianpro,zhang2024pixel}. In this paper, we adopt the Gaussian Opacity Fields model \cite{ZehaoTOG24GaussianOpacityFields} as our 
3D representation, which integrates the anti-aliasing module from \cite{yu2024mip} and offers highly accurate geometry through regularization, critical for visual localization. 

\PAR{Camera pose refinement-based visual localization} 
Pose refinement approaches 
minimize, with respect to the pose, the difference between the query image and a rendering obtained by projecting the scene representation into the image using the current pose estimate~\cite{alismail2017photometric,engel2014lsd,EngelPAMI17DirectSparseOdometry,schopscvpr19DADSLAM,SchopsCVPR17MultiViewStereo}. The pose is iteratively refined from an initial  estimate.
L2 distances  between deep features are often used to measure the differences between the query image and the projected features from the 
scene~\cite{von2020gn,vonstumberg2020lmreloc,SarlinCVPR21BackToTheFeature,GermainCWPRWS21FeatureQueryNetworks,lindenberger2021pixelperfect,xu2020deep}. 
\cite{trivigno2024unreasonable} integrates pre-trained features with a particle filter, thus removing the need to learn representations per scene. 
\cite{pietrantoni2023segloc} replaces deep features with segmentation labels and aligns them with 2D segmentation maps. 
In contrast, our \gsf{} Pose Refinement pipeline (\gsfloc{}) 
associates a deep feature or a segmentation label to each 3D Gaussian primitive and performs pose refinement by 1) rasterizing these quantities and 2) explicitly backpropagating the feature-metric or segmentation errors through the rasterizer with respect to the camera pose. 

\PAR{Rendering-based visual localization}
Leveraging the capability of novel view synthesis of NeRFs, iNeRF~\cite{yenchen2021inerf,lin2023parallel} is the first to 
perform pose refinement through photometric alignment, where the optimization is performed by direct gradient descent back-propagation through the neural field. \cite{maggio2023loc,lin2023parallel,trivigno2024unreasonable} parallelize this process with faster neural radiance fields or particle filters. 
NeRF-based representations have used to match \cite{MoreauX23CROSSFIRECameraRelocImplicitRepresentation,zhou2024nerfect}  or align rendered features from an implicit field~\cite{chen2023refinement,pietrantoni2024jointnerf}.

Multiple visual localization works have emerged leveraging the efficient 3DGS models instead of NeRFs. 
\cite{bortolon20246dgs} matches rays emerging from ellipsoids with pixels with an attention mechanism and estimates the pose with a closed form solution. 
\cite{liu2024gsloc}  performs matching between query and 
rendered image, lifting the matches in 3D with the rendered depth to refine the pose by solving a PnP problem. Similar to NeRF-based approaches, \cite{sidorov2024gsplatloc,zhai2024splatloc,zhai2025neuraloc} distills pretrained features in a Gaussian-based 3D feature field.  
These methods estimate poses by matching these features followed by PnP+RANSAC.
In contrast,
we do not rely on pretrained features but learn our representations in a self-supervised manner. Furthermore, we solely estimate query poses through feature-metric refinement by explicitly backpropagating through the rasterizer, which yields higher accuracy even without additional  PnP+RANSAC step.
Furthermore, contrarily to concurrent works our 3D features take scale into account.

\PAR{Privacy-preserving visual localization} 
\citet{pittaluga2019revealing} have shown that detailed and recognizable images of a scene can be obtained from sparse 3D point clouds associated with local descriptors.  While, 
\cite{SpecialeCVPR19PrivacyPreservingImageBasedLocalization,SpecialeICCV19PrivacyPreservingImageQueriesforCameraLocalization,ShibuyaECCV20PrivacyPreservingVisualSLAM} tried to address this by 
transforming 3D point clouds into 3D line clouds, it was shown in \cite{ChelaniCVPR21HowPrivacyPreservingAreLineClouds,lee2023paired} that
these line clouds still preserve a significant amount of information about the scene geometry that can be used to recover 3D point clouds, hence enabling the inversion attack~\cite{pittaluga2019revealing}. Instead,
\cite{lee2023paired} lift pairs of points to lines and \cite{moon2024efficient}
relies on a few spatial anchors resulting in 3D lines with non uniformly sampled distributions.
Another 
strategy is to swap coordinates between random 
pairs of points \cite{pan2023privacy} or performing partial pose estimation against distributed partials maps \cite{geppert2022privacy}, but as shown in  \cite{chelani2024obfuscation}, the
point positions may be recovered by identifying points neighborhoods from co-occuring descriptors.

 Another line of work studies the privacy on the feature representation level.  \cite{DusmanuCVPR21PrivacyPreservingImageFeaturesViaAdversarialAffineSubspaceEmbeddings,pittaluga2023ldp} embed descriptors in 
 subspaces designed to withstand different privacy attacks.
 Closer to our work, \cite{zhou2022geometry,wang2023dgc,pietrantoni2023segloc} remove the need for high-dimensional features. 
 \cite{zhou2022geometry,wang2023dgc} match keypoints and 3D points without descriptors and \cite{pietrantoni2023segloc} replace high-dimensional descriptors with segmentation labels. 
Similar to \cite{pietrantoni2023segloc}, we also rely on labels to ensure privacy, but
our dense label rendering through 3DGS 
achieves better localization accuracy
than \cite{pietrantoni2023segloc}, which  relies on reprojected sparse labels.

\begin{figure*}[tt]
\centering
\includegraphics[width=0.95\linewidth]{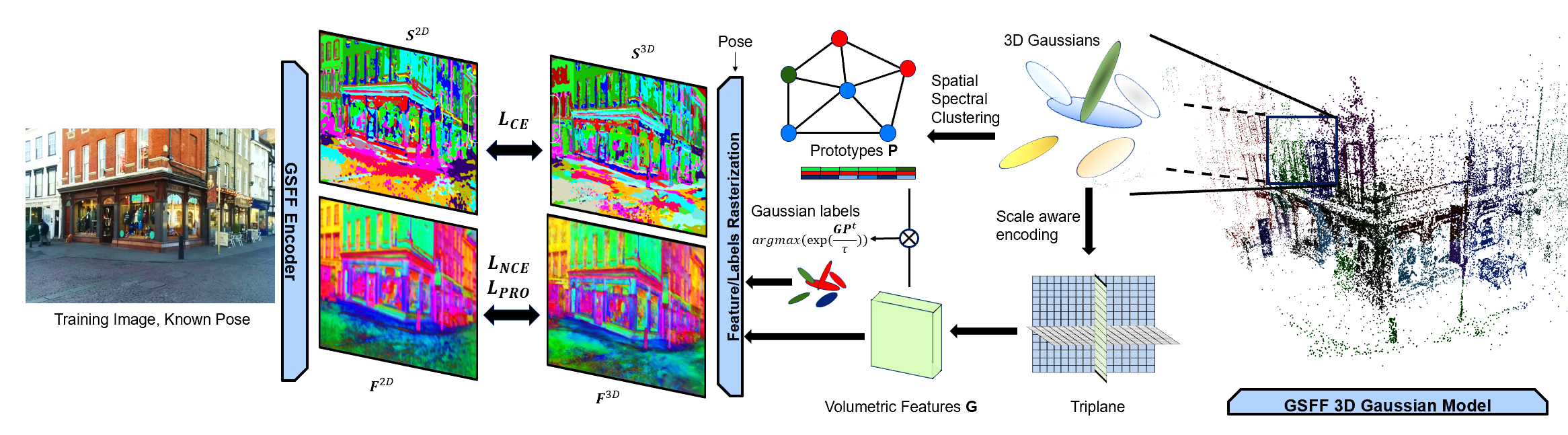}
\caption{\textbf{Training pipeline}. We extract a feature map $\bfF^\text{2D}$ and segmentation map $\bfS^\text{2D}$ from a training image with known pose (left). For each 3D Gaussian $\calG_i$ (right), a scale aware feature $\bfg_i$ is extracted from a triplane representation. Spectral clustering is applied on the Delaunay graph derived from the Gaussian cloud, yielding a set of prototypes $P$. A label is associated with each 3D Gaussian by assigning the volumetric features to the prototypes. The features and labels are then rendered to obtain the feature map $\bfF^\text{3D}$ and the segmentation map $\bfS^\text{3D}$, which are respectively aligned with their encoder counterparts $\bfF^\text{2D}$ and $\bfS^\text{2D}$  through $L_{NCE},L_{PRO}$ and $L_{CE}$.}
\label{fig:main_fig}
\vspace{-.5cm}
\end{figure*}

\section{3DGS Feature Fields (\gsf{})}
\label{sec:GSF2PR}

First in \cref{sec:feature_field} 
we introduce our Gaussian Splatting Feature Field (\gsf) which associates a scale aware 
feature to each 3D Gaussian. We describe how to 
jointly optimize it along with a 2D feature extractor by aligning 
rendered and 2D extracted features in a self-supervised manner.  In \cref{sec:prototypes}
we propose to further improve the features' discriminativeness and their 3D awareness, 
by clustering the 3D Gaussian cloud and encouraging that both features in a pixel aligned pair of 2D/3D features are close to the same prototype through an auxiliary contrastive loss.
These prototypes not only help distilling spatial information in the feature space, but also allow for a smooth transition from feature maps to segmentation maps, 
enabling privacy-preserving localization 
(as described in \cref{sec:PPGSF4Loc}). 
Finally, in \cref{sec:loc_pip} we describe the \textit{\gsfloc{} Feature} visual localization pipeline 
that is based on pose refinement through 
feature alignment.  
Its extension, the privacy-preserving \textit{\gsfloc{} Privacy} visual localization pipeline 
based on segmentations is described in \cref{sec:PPGSF4Loc}.
We show an overview of the architecture and training pipeline in Fig.~\ref{fig:main_fig}.

\subsection{Scene Representation}
\label{sec:feature_field}

We adopt the Gaussian Opacity Fields model \cite{ZehaoTOG24GaussianOpacityFields} as our 
Gaussian Field representation,
where a scene is composed of a set of 3D Gaussian primitives parametrized by their center,  scaling and rotation  matrices, opacity  and spherical harmonics coefficients. 
Given a ray $r$ emanating from a camera pose $P$, \cite{ZehaoTOG24GaussianOpacityFields} finds the intersection between the ray and 3D Gaussians and computes the contribution $C_i(r,P)$ of each Gaussian $\calG_i$ traversed by the ray. 
The Gaussians are then ordered based on depth and the pixel's color is obtained by alpha blending 
\begin{align}
    c(r,P) = \sum_{i=1}^N c_i \alpha_i C_i(r,P) \prod_{j=1}^{i-1} (1-\alpha_j C_j(r,P)) \enspace , \label{eq:blending}
\end{align}
where $c_i$ is the view-dependent color, modeled with spherical harmonics associated with $\calG_i$ and $\alpha_i$ are the blending weights. 
The ray-Gaussian intersection formulation allows for depth distortion and normal consistency regularization \cite{ZehaoTOG24GaussianOpacityFields},  which improves the geometry of the 3D scene. 

\myparagraph{3D feature fields} 
Similar to \cref{eq:blending}, features or segmentation labels can be 
 rendered through the same alpha blending  by simply replacing each Gaussian's color by the feature or segmentation label assigned to the Gaussian.  However, this requires that we assign to each Gaussian $\calG_i$ a  feature, which in general might be extremely costly, especially for
 high-dimensional features. To avoid this, instead of associating independent features to each Gaussian, we introduce a feature field parametrized by a triplane grid \cite{chen2022tensorf}.
It is centered at the origin of the world coordinate space and is composed of three two-dimensional orthogonal planes $H_{xy},H_{xz},H_{yz} \in \Re^{R \times R}$, 
where $R$ is the resolution of the grid.  
For simplicity, we will also denote by $H_{xy},H_{xz},H_{yz} \in \Re^{R \times R \times D}$  (and use them  interchangeably) the three corresponding feature tensors, where $D$ is the dimensionality of the 
feature space we aim to learn. 

To compute the volumetric feature $\bfg_i$ associated with a 3D Gaussian $\calG_i$ from the triplane, we proceed as follows.  We 
project the 3D Gaussian $\calG_i$ onto the three planes 
and compute for 
the three resulting 2D Gaussians $\calG^{xy}_i,\calG^{xz}_i,\calG^{yz}_i$ 
three corresponding features $\bfg^{xy}_i,\bfg^{xz}_i,\bfg^{yz}_i$ by applying an RBF kernel parametrized by the 2D Gaussians
(as  detailed in the \supmat).
The three grid features are averaged yielding the volumetric
feature $\bfg^\text{3D}_i$ associated with the Gaussian $\calG_i$.  
This representation, 
called  Gaussian Splatting Feature Field (\gsf), enables the features to be scale-aware, as 3D Gaussians with large spatial span will aggregate feature information over a large area within the grid,  
and small Gaussians over 
smaller areas. 
Further advantages of the \gsf{} representation are:  1) they allow
sharing information between Gaussians based on overlapping projections onto the planes as the Gaussian features are  optimized interdependently through rasterization, 
and 2)  the field can be queried from any 3D position 
making it suitable for 3DGS splitting and merging mechanisms.

\myparagraph{Self-supervised training}
Our aim is to perform pose refinement
based on aligning pixel-level 2D encoded features from the image with 3D rendered features. Hence it is important for these feature to be locally discriminative and robust to viewpoint changes. 
Formally, we want to learn the  \gsf{} feature field jointly with a 2D encoder such that when we render, at pose $P$, a feature map $\bfF^\text{3D}$ from the feature field, it is aligned with 
the corresponding 2D feature map $\bfF^\text{2D}$ 
of the image $I$ associated to the pose $P$. 
The  rendered feature map 
$\bfF^\text{3D}$ is obtained with alpha blending similarly
to \cref{eq:blending}, where for each pixel $u$,
we replace $c_i$ with $\bfg^\text{3D}_i$. 
To align 
$\bfF^\text{3D}$ and $\bfF^\text{2D}$, we train the model in a self-supervised manner with the contrastive loss
\cite{oord2018representation}:
\begin{align}
L_{NCE} = -\frac{1}{2HW}  \sum_{u \in I}  \log \left( \frac{\exp{\left(\bfF^\text{3D}_{u} \cdot \bfF^\text{2D}_{u}/\tau \right)}^2}{A} \right), 
 \label{eq:nce}
\end{align}
where $\tau$ is the temperature parameter, 
and $A$ is a normalizing factor (detailed in the \supmat).

\begin{figure*}[t]
\centering
\includegraphics[width=1\linewidth]{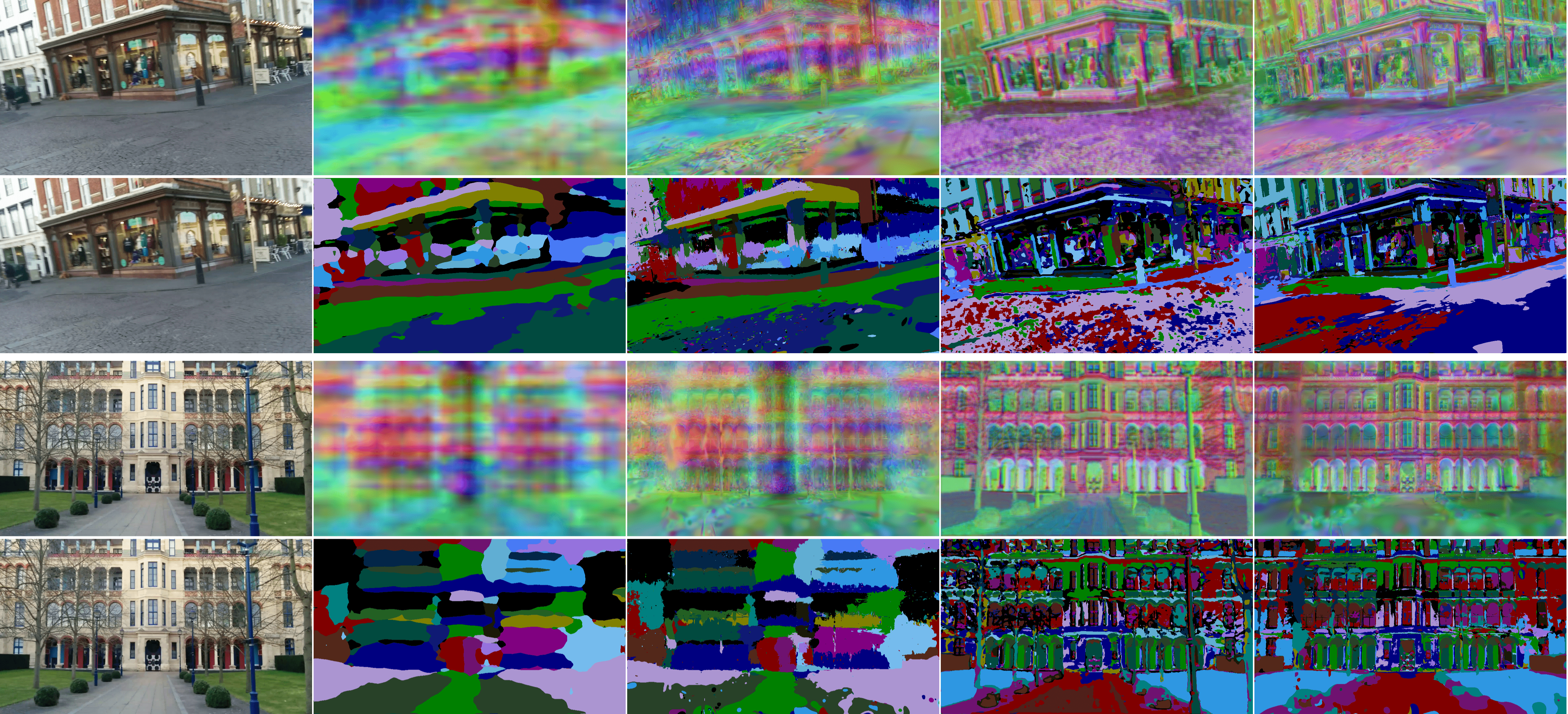}
\caption{First and third lines:  original image, encoder feature map  $\bfF^\text{2D}$
(second and forth) and rendered feature map 
$\bfF^\text{3D}$ (third and fifth) at  coarse level (second and third)  and fine level (forth and fifth).
Below (second and forth lines) correspondingly  rendered image, 
encoder segmentation maps  $\bfS^\text{2D}$
(second and forth) and rendered segmentation  maps
$\bfS^\text{3D}$  (third and fifth) at  coarse and fine level respectively. 
Features are visualized through PCA downprojection. }
\label{fig:visu_segs}
\end{figure*}

\subsection{Prototypical Feature Regularization}
\label{sec:prototypes}

To better align the features and prepare the transition from features to segmentations, we 
propose to structure the feature space around a set of classes. To that end, we cluster the feature field, derive  3D prototypes, and apply an auxiliary contrastive loss to enforce that corresponding pixel-aligned 2D extracted  and  3D rendered features  are close to the same prototypes in the feature space.

\myparagraph{Spatial prototypes} 
We want to find  a set of feature prototypes that best encode the spatial prior reflecting the Gaussian cloud structure. One option would be to apply spectral clustering on a matrix containing pairwise distance between Gaussian centers. However given the high number of Gaussians, this would quickly become untractable. Therefore, we start with applying a Delaunay triangulation \cite{delaunay1934sphere} of the Gaussian centers, which yields a graph that already captures local geometric information.  Furthermore, the eigenvalues  and eigenvectors can now be computed from the Laplacian of the sparse adjacency matrix derived from the graph.  
The resulting eigenvectors, one for each Gaussian $\calG$, are clustered into $K$ groups, which assigns each 3D Gaussian $\calG_i$ to a cluster $k$. 
 To  build the representative feature (\textit{cluster prototype}) $\bfp_k$ for  each cluster, we simply average the volumetric features $\bfg_i$
of all Gaussians  belonging to cluster $k$.

\myparagraph{Prototypical loss} These prototypes implicitly define "classes" and we use them to maximize intra-class compactness and inter-class separability within the feature space and infuse spatial priors. Given a batch of N pairs of pixel-aligned rendered/encoder features $\{\bfF^\text{3D}_u,\bfF^\text{2D}_u\}$, we associate one prototype $\bfp_k$ per feature pair and want to enforce both features to be close to the prototype. 
We thus add  the following prototypical contrastive loss: 

\vspace{-0.3cm}
{\small
\begin{align}
L_{PRO} = -\frac{1}{N} \sum_{n=1}^N \log\left(\frac{
 \exp{(\bfF^\text{3D}_{n} \bfp_{n}^t +\bfF^\text{2D}_{n} \bfp_{n}^t)/\tau )}}{B}\right) ,
 \label{eq:pro}
\end{align}}
where $\bfF^\text{2D}_{n},\bfF^\text{3D}_{n}$ are the features 
corresponding to the  pixel $u_n$, $\bfp_{n}$ is the prototype assigned to them, $\tau$ is a temperature parameter
and $B$  is  a normalizing factor  (detailed in the \supmat{} along with derivations). To compute the associations over the batch of feature pairs, we use an optimal transport procedure based on the Sinkhorn-Knopp algorithm~\cite{distances2013lightspeed} (see the \supmat{}  for details). 

\myparagraph{Multi-view consistency}
The rendering depends on the incidence of the ray traversing the 3D Gaussians. Hence rendered features may vary across viewing angles. 
In order to encourage multi-view consistency of both the encoder and the feature field, and to make our features generalizable   to out-of-distribution views, 
we align features  belonging to different views but  associated to the same 3D points.
To that end, given a pixel correspondence $(u,v)$ between two images $I$ and $\widehat{I}$ in the scene -- along with their encoder/rendered features $\{\bfF^\text{3D},\bfF^\text{2D}\}$, $\{\widehat{\bfF}^\text{3D},\widehat{\bfF}^\text{2D}\}$ --  we randomly replace in $L_{NCE}$ and $L_{PRO}$ a subset of the pixel aligned feature pairs $\{\bfF^\text{3D}_u,\bfF^\text{2D}_u\}$ by $\{\widehat{\bfF}^\text{3D}_v,\bfF^\text{2D}_u\}$ or $\{\bfF^\text{3D}_u,\widehat{\bfF}^\text{2D}_v\}$.
To extract correspondences, given a training image $I$ with pose $P$ and rendered depth $D$, we generate a random pose $\widehat{P}$ in the vicinity of $P$. We render the image $\widehat{I}$, depth $\widehat{D}$, and features $\widehat{\bfF}^\text{3D}$ from pose $\widehat{P}$ and extract the 2D feature map  $\widehat{\bfF}^\text{2D}$ from the rendered image.  
Then, for a randomly sampled set of pixels $u$ in $I$, we backproject it using $D$ and re-project it into $\widehat{I}$, yielding the pixel $v$ in the rendered image. Then,  we backproject $v$ into 3D using $\widehat{D}$  and re-project it into $I$. If the pixel distance between $u$ and the reprojected $v$ fall within a small threshold, $(u,v)$ is considered as a valid correspondence, otherwise it is rejected.

\begin{table*}[tt]
\scriptsize
\resizebox{\linewidth}{!}{
  \begin{tabular}{cl||c|c|c|c|c|c|c}
   &  Model  & Chess & Fire & Heads & Office & Pumpkin & Redkitchen & Stairs  \\ 
   \hline
 \multirow{4}*{\begin{sideways} SBM \end{sideways}}  
 & HLoc~\cite{SarlinCVPR19FromCoarsetoFineHierarchicalLocalization} & 0.8/0.11/100 & 0.9/0.24/99 &  0.6/0.25/100 & 1.2/0.20/100 & 1.4/0.15/100 & 1.1/0.14/99  &2.9/0.80/72 \\ 
   & DSAC*\cite{brachmann2021visual} & 0.5/0.17/100 &  0.8/0.28/99 & 0.5/0.34/100 & 1.2/0.34/98 & 1.2/0.28/99 & 0.7/0.21/97 & 2.7/0.78/92 \\ 
  & ACE \cite{brachmann2023accelerated} & 0.5/0.18 & 0.8/0.33 & 0.5/0.33 & 1.0/0.29 & 1/0.22 & 0.8/0.2 & 2.9/0.81 \\ 
  & ACE + GS-CPR \cite{liu2024gsloc} & 0.5/0.15 & 0.6/0.25 & 0.4/0.28 & 0.9/0.26 & 1.0/0.23 & 0.7/0.17 & 1.4/0.42 \\ 
  \hline
  \multirow{7}*{\begin{sideways} RBM  \end{sideways}}
   & NeFeS~\cite{chen2023refinement} (DFNet \cite{ShuaiECCV22DFNetEnhanceAPRDirectFeatureMatching}) &  2/0.79 & 2/0.78 & 2/1.36 & 2/0.60 & 2/0.63 & 2/0.62 & \underline{5/1.31} \\ 
   & MCLoc \cite{trivigno2024unreasonable} &  2/0.8 & 3/1.4  & 3/1.3 & 4/1.3  & 5/1.6 & 6/1.6 & 6/2.0  \\
&     SSL-Nif ~\cite{pietrantoni2024jointnerf} (DV)&  1/0.22/93 & 0.8/0.28/91 & 0.8/\underline{0.49}/71 & 1.7/0.41/81 & \underline{1.5}/\underline{0.34}/86 & 2.1/0.41/75 & 6.5/0.63/49 \\ 
 & NeRFMatch ~\cite{zhou2024nerfect} (DV) &  0.9/0.3 & 1.3/0.4 & 1.6/1.0 & 3.3/0.7 & 3.2/0.6 & 1.3/0.3 & 7.2/1.3 \\ 
& GSplatLoc \cite{sidorov2024gsplatloc} (GSplatLoc)&   \bf0.43/0.16 & 1.03/0.32 & 1.06/0.62 & 1.85/0.4 & 1.8/0.35 & 2.71/0.55 & 8.83/2.34 \\ 
& GS-CPR \cite{liu2024gsloc} (DFNet)& 0.7/0.20 & \underline{0.9/0.32} & \underline{0.6}/\textbf{0.36} & \underline{1.2/0.32} & \bf1.3/0.31 & \underline{0.9/0.25} &  \textbf{2.2/0.61} \\ 
& \textbf{GSFFs-PR} Feature (DV) &  \textbf{0.4}/\underline{0.19}/95 & \textbf{0.6}/\textbf{0.26}/98 & \textbf{0.5/0.36}/94 & \textbf{1.0}/\textbf{0.31}/97 & \textbf{1.3}/\underline{0.38}/85 & \bf{0.6}/\bf{0.23}/93 & 25.1/0.63/32 \\
\hline
& \textbf{GSFFs-PR} Privacy (DV) &   
 0.8/0.28/96 & 0.8/0.33/94 & 1.0/0.67/90 & 1.5/0.51/90 & 2.0/0.50/78 & 1.2/0.33/85 & 28.2/0.98/29 \\
     \hline
     \bottomrule
    \end{tabular}
}
\caption{\textbf{Localization results on 7Scenes} using SfM-based pseudo GT poses from \cite{BrachmannICCV21OnTheLimitsPseudoGTVisReLoc}.  We compare our results with Structure-Based Methods (SBM) and 
 Rendering-Based Methods (RBM). Median position error (cm.) ($\downarrow$)/ median rotation error (°) ($\downarrow$)/ recall at 5cm/5° ($\%$) ($\uparrow)$, when available. 
 \textbf{Bold} are the best results, \underline{underline} the second best.
 }
      \label{tab:7scenesWithpGT}
    \end{table*}

\subsection{Feature-based Localization (\gsfloc)}
\label{sec:loc_pip}
After training, the feature space along with the 3D Gaussians can be used for pose refinement by finding the pose that minimizes feature-metric errors between 2D extracted query features and 3D 
features rendered from the aforementioned pose.
 In particular, given a query image whose pose is unknown, we first extract its 2D feature map  $\bfF^\text{2D}$ with our trained encoder. Second, we retrieve the closest database image with a global descriptor (our method is agnostic to the global descriptor used) and use the pose $P_{init}$ of the retrieved image as initialization. From $P_{init}$, we render the  $\bfF^\text{3D}$ feature map 
 from   \gsf{}. 
 Feature inconsistencies between the two feature maps are iteratively minimized with regard to the pose:
 \begin{align}
 P^*=\text{min}_{P\in SE(3)} \Vert \bfF^\text{2D} - \bfF^\text{3D}(P,\calG) \Vert_2^2 \enspace,
 \label{eq:featalign}
 \end{align}
where the pose is parametrized on SE(3) and updates are done on the Lie algebra se(3) by backpropagating explicitly through the rasterizer \cite{matsuki2024gaussian} (that we modify to feature-metric refinement). At each refinement iteration, the pose is updated and features are rendered from this updated pose. We  minimize the \cref{eq:featalign} objective until convergence.

\section{Privacy-Preserving \gsf{}}
\label{sec:PPGSF4Loc}

Due to the clustering and subsequent prototypical formulation adopted in \gsf, we can  seamlessly  extend our model to be privacy-preserving. 
We follow the privacy definition from the literature \cite{zhou2022geometry,pietrantoni2023segloc,wang2023dgc} where privacy is refers to the inability for an attackers to recover texture/color information and fine-level details. Similarly, following prior works, we consider that having coarse geometric information does not violate privacy. 
Therefore,  
inspired by \cite{pietrantoni2023segloc}, 
we replace  the
high-dimensional features $\bfg_i$ that contain a lot of privacy-critical information with a single segmentation label. 
After training, the feature field, prototypes, and color information are removed.

\myparagraph{Optimizing the segmentations}
Features may be converted to segmentations by soft assigning them to the set of prototypes.
Any Gaussian feature $\bfg_i$ or encoder feature $\bff_i$ can be assigned to 
a prototype $\bfp_k$ by the
likelihood  of the feature belonging to the cluster $k$ defined by $l^{3D}_{ik}= \exp(\bfg_i^\top \bfp_k)/\sum_{k'} \exp(\bfg_i^\top \bfp_{k'})$. 
The resulting scores form a pseudo-logits vector $\bfl_i$ that can be rendered with alpha-blending  (\cf  \cref{eq:blending}), where the color is replaced by the scores.
Similarly, encoder features can be soft-assigned via $l^{2D}_{ik}=\exp(\bff_i^\top \bfp_k)/\sum_{k'} \exp(\bff_i^\top \bfp_{k'})$.  

While the prototypical learning framework induces relatively well aligned encoder/rendered soft assignments, the localization accuracy can  further be improved by learning to directly predict the 2D segmentations.  
Therefore, during training, 
we add and train a shallow classification head on top of the encoded features that learns to output the segmentation map $\bfS^\text{2D}$ directly from the images. 
To learn the segmentation head and to further  refine our features in a self-supervised way, we  add the following cross-entropy loss to \cref{eq:nce} and \cref{eq:pro}:
\begin{align}
    L_{CE} = - \sum_{u \in I} \bbone_u
    \cdot \left(\log(\bfS^\text{2D}_{u})  + \log(\bfS^\text{3D}_{u})\right) \enspace ,
\label{eq:ce}
\end{align}
where $\bfS^\text{3D}$ denotes the segmentation map with rendered pseudo-logits   $l^{3D}_{ik}$, $\bbone_u$ is the one hot vector corresponding to the prototype label associated to the pixel $u$, using 
the  exact same associations as in the prototypical contrastive loss  encouraging consistency between the features and segmentations.

\myparagraph{Privacy-preserving localization, \textbf{\gsfloc} Privacy}
\label{sec:loc_pip_pri}
After training, any potentially privacy-sensitive information is removed from the 3D Gaussian model (spherical harmonic, features, and \gsf) as neither photometric information nor features are required for the privacy-preserving localization. 
The pseudo-logits obtained from the soft-assignments still contain too much information \cite{pietrantoni2023segloc}. 
Therefore, we instead store hard assignments in the form of a single label per Gaussian $\calG_i$, $k^\ast = \text{argmax}_k(\bfl_{ik})$. After this labeling procedure, the triplane \gsf{} feature field and the prototypes are subsequently removed. As such the 3D models contain only geometric (Gaussians without color information) and  segmentation information (cluster labels),   effectively increasing the level of privacy.  
Furthermore, storing only the cluster labels  makes the storage of the 3D representation orders of magnitudes smaller  compared to storing features. 
Given a pose $P$, segmentation maps can be 
rendered by alpha-blending (via \cref{eq:blending}) from  one-hot vectors $\bbone_{k}$ (corresponding to label $k$), 
yielding $\bfS^\text{3D}_{P}$. Given an image,  2D segmentation maps $\bfS^\text{2D}$ are directly obtained through the segmentation head.
The localization pipeline is similar to the one in \cref{sec:loc_pip} with the difference that we minimize segmentation inconsistencies between the segmentation labels instead of features:
 \begin{align}
 P^*=\text{min}_{P\in SE(3)} CE(\bfS^\text{2D},\bfS^\text{3D}_{P}) \enspace . 
 \label{eq:featalignpp}
 \end{align}

\section{Experimental Evaluation}
In this section, we first provide details regarding the training and evaluation of \gsfloc. Then, in \cref{sec:vl} we evaluate our feature-based VL pipeline (dubbed \textbf{\gsfloc} Feature) and our privacy-preserving VL pipeline (dubbed \textbf{\gsfloc} Privacy)  on multiple real world datasets. We show that \gsfloc{} in general is more accurate than the respective non-privacy and privacy-preserving baselines. 
\cref{sec:abl} ablates and discusses important aspects of our approach.

\begin{table*}[tt]
  \centering
   \resizebox{0.8\linewidth}{!}{
  \begin{tabular}{cl||c|c|c|c}
   &  Model  & King's & Old & Shop & St. Mary's \\ 
    \hline
     \multirow{5}*{\begin{sideways}  SBM \end{sideways}}
    & HLoc~\cite{SarlinCVPR19FromCoarsetoFineHierarchicalLocalization}  & 11 / 0.20 & 15 / 0.31 & 4 / 0.2 & 7 / 0.24 \\ 
     & Kapture R2D2 \cite{humenberger2020robuste}  & 5 / 0.10 & 9 / 0.20 & 2 / 0.10 & 3 / 0.10 \\ 
    & DSAC*\cite{brachmann2021visual} & 15 / 0.30 & 21 / 0.40 & 5 / 0.30 & 13 / 0.40 \\ 
    & ACE \cite{brachmann2023accelerated} & 29 / 0.38 & 31 / 0.61 & 5 / 0.30 & 19 / 0.60 \\ 
    & ACE + GS-CPR \cite{liu2024gsloc} & 25 / 0.29 & 26 / 0.38 &5 / 0.23 & 13 / 0.41 \\
   \hline
   \multirow{4}*{\begin{sideways} PPM \end{sideways}} 
   & GoMatch \cite{zhou2022geometry} & 25 / 0.64 & 283 / 8.14 & 48 / 4.77 & 335 / 9.94 \\ 
   &  DGC-GNN \cite{wang2023dgc}  &  \textbf{18} / 0.47 & 75 / 2.83 & 15 / 1.57 & 106 / 4.03 \\ 
   & SegLoc \cite{pietrantoni2023segloc} (DV) & \underline{24} / \bf{0.26} & \underline{36 / 0.52} & \underline{11 / 0.34} & 17 / \textbf{0.46} \\ 
   & \textbf{GSFFs-PR} Privacy (DV) & \underline{24} / \underline{0.39} & \bf{26 / 0.49} & \bf{5 / 0.27}
   & \textbf{13} / \underline{0.48} \\
 \hline
     \multirow{8}*{\begin{sideways} RBM \end{sideways}}
 &   CROSSFIRE~\cite{MoreauX23CROSSFIRECameraRelocImplicitRepresentation} (DV)  & 47 / 0.7 & 43 / 0.7 & 20 / 1.2 & 39 / 1.4 \\ 
   & NeFeS~\cite{chen2023refinement} (DFNet) & 37 / 0.62 & 55 / 0.9 & 14 / 0.47 & 32 / 0.99 \\ 
  & MCLoc \cite{trivigno2024unreasonable} & 31 / 0.42 & 39 / 0.73 & 12 / 0.45 & 26 / 0.88  \\
    &  SSL-Nif \cite{pietrantoni2024jointnerf} (DV) & 32 / 0.40 & 30 / 0.49 & 8 / 0.36 & 24 / 0.66 \\ 
     & NeRFMatch ~\cite{zhou2024nerfect} (DV) &  \bf{13} / \bf{0.2} & 21 / \underline{0.4} & 8.7 / 0.4 & 11.3 / 0.4  \\

     &  GS-CPR\cite{liu2024gsloc} (DFNet) & 23 / 0.32 &  42 / 0.74  &10 / 0.36 &27 / 0.62 \\
     
      & GSplatLoc \cite{sidorov2024gsplatloc} & 27 / 0.46 & \underline{20} / 0.71 & \underline{5} / 0.36 & 16 / 0.61 \\
     & \textbf{GSFFs-PR} Feature (DV)  & 18 / 0.27 & 21 / \underline{0.4} & \textbf{4} / \underline{0.26} & \underline{10} / \underline{0.34}  \\
    & \textbf{GSFFs-PR} Feature (tuned) (DV)  & \underline{17} / \underline{0.26} & \bf{18} / \bf{0.36} & \bf{4} / \bf{0.25} & \bf{8} / \bf{0.26}  \\ 
    
    \hline
    \bottomrule
    \end{tabular}
    }
    \caption{Localization results on Cambridge Landmarks. Comparison with Rendering-Based Methods (RBM),  Privacy-Preserving Methods (PPM) and Structure-Based Methods (SBM). Median position error (cm.) ($\downarrow$), median rotation error (°) ($\downarrow$). \textbf{Bold}: 
    best results per method category, \underline{underline}: 
    second best. \textit{(tuned)} means that training hyperparameters are tuned for outdoor dataset as opposed to indoor dataset.
     }
      \label{tab:cambridge}
    \end{table*}

\PAR{Experimental details}
We train and optimize one model per scene for 50K iterations on a single NVIDIA A100 GPU. During the first 15K iterations only the Gaussians parameters are updated through the photometric loss. From iteration 15k, the triplane fields and the encoder are also optimized with the joint loss $L = L_{PHO} + .5 L_{NCE} + .5 L_{PRO} + .5 L_{CE} + .1 L_{TVL} + 0.05 L_{Depth}$, 
where $L_{TVL}$ is a total variation loss applied to the triplane parameters for regularization and $L_{Depth}$ is a depth prior when available.  To increase the convergence basin of our method, we define a coarse feature/segmentation space
and a fine feature/segmentation space (both are trained in a similar fashion).  The coarse triplane has a resolution of $R=256$ and the fine triplane  a resolution of $R=1024$. Coarse image-based features are processed through a ViT \cite{oquab2023dinov2} with patch size of 14, while fine image-based features are not downsampled. The feature dimension $d$ is set to 16 for both fine and coarse levels and the number of cluster $K$ to 34 both for fine and coarse levels. A query image estimated pose is first iteratively refined with the coarse-level features or segmentations, then it is refined with the fine-level.

We evaluate our localization pipeline on the 7Scenes (with  Depth SLAM pseudo ground truth poses \cite{ShottonCVPR13SceneCoordinateRegression} and 
with SfM-based pseudo ground truth poses~\cite{BrachmannICCV21OnTheLimitsPseudoGTVisReLoc}), Indoor6 \cite{do2022learning}, 
and  Cambridge Landmarks~\cite{KendallICCV15PoseNetCameraRelocalization} real-world datasets. Results on the 12Scenes \cite{valentin2016learning} dataset and comparisons against \cite{zhai2024splatloc,zhai2025neuraloc} are 
provided in the \supmat. 
The median position error (cm) and  median rotation error (°) are reported. Additionally, we report the recall at 5cm/5° for the indoor scenes. 
Training parameters (loss coefficients, temperature, 3DGS hyperparameters) are kept the same for all datasets. More details are provided in the \supmat.

 \subsection{Visual Localization}
\label{sec:vl}
\myparagraph{Comparison with feature rendering-based approaches}
We primarily compare \gsfloc{} Feature 
against other non-privacy-preserving rendering-based VL methods, which either use NeRFs \cite{chen2023refinement,pietrantoni2024jointnerf,zhou2024nerfect} or 3DGS 
\cite{liu2024gsloc,sidorov2024gsplatloc} as their underlying 3D representations. These methods all require an initial pose estimate, which is obtained via pose retrieval with a global descriptors. More accurate global descriptors lead to more accurate initial pose estimates, which in turn makes refinement easier. For the sake of fairness, we try to use similar 
or less accurate global descriptors as the other baselines.  
The global descriptor used is specified between brackets. 
We report 7Scenes results (with SfM pseudo GT) 
in \cref{tab:7scenesWithpGT}, and Cambridge Landmarks results in \cref{tab:cambridge}.

From \cref{tab:7scenesWithpGT}, we observe that our \textbf{GSFFs-PR} Feature (DV) is best on 4 out of 7 scenes and second for two further ones in spite of relying on a less   accurate initialization, \ie,   DenseVLAD 
is less accurate than DFNet or ACE.  On 
 Stairs, the Opacity Gaussian Field struggles
 due to its textureless and flat layout, 
 yielding a poor results for the initial 3D structure that directly affects  
 \gsfloc's refinement framework.   
From \cref{tab:cambridge}, 
we can see that \gsfloc{} is more accurate than other 3D Gaussian rendering-based baselines. Overall, our feature pipeline even displays competitive performances on some scenes compared to state-of-the-art structure-based methods.

     \begin{table*}[tt]
\centering
  \begin{tabular}{l||c|c|c|c|c|c|c}
     Model  & Chess & Fire & Heads & Office & Pumpkin & Redkitchen & Stairs  \\ 
   \hline  
    GoMatch \cite{zhou2022geometry} &  4/1.65 & 13/3.86 & 9/5.17 & 11/2.48 & 16/3.32 & 13/2.84 & 89/21.12  \\
     DGC-GNN \cite{wang2023dgc} &  3/1.43 & 5/1.77 & 4/2.95 & 6/1.61 & 8/1.93 & 8/2.09 & 71/19.5  \\
    SegLoc \cite{pietrantoni2023segloc} &  4.5/1.02 & 4.0/1.51 & 4.1/1.80 & 4.7/1.33 & 6.3/1.88 & 6.0/1.87 & 43.7/8.27  \\
    \textbf{GSFFs-PR}  & \bf2.9/0.92 & \bf2.3/0.91 & \bf1.5/1.11 & \bf4.0/1.28 & \bf5.7/1.48 & \bf5.1/1.66 & \bf28.3/2.39 \\
     \hline
     \bottomrule
    \end{tabular}
\caption{Localization results on 7Scenes (DSLAM based pseudo GT), with median position error (cm.) ($\downarrow$)/ median rotation error (°) ($\downarrow$).
We compare \textbf{GSFFs-PR} (Privacy)
with other privacy-preserving methods. Both SegLoc and \textbf{GSFFs-PR} refine the pose from  DenseVLAD \cite{ToriiPAMI18247PlaceRecognitionViewSynthesis} initialization.
 Best numbers in \textbf{bold}.}
      \label{tab:7scenesWithDslam}
    \end{table*}

\begin{table*}[tt!]
\scriptsize
\resizebox{\linewidth}{!}{
  \begin{tabular}{cl||c|c|c|c|c|c}
     &Model & scene1 & scene2a & scene3 & scene4a & scene5 & scene6  \\
    \hline
     \multirow{4}*{\begin{sideways} NPPM \end{sideways}}  
    & DSAC* \cite{brachmann2021visual} & 12.3/2.06/18.7 & 7.9/0.9/28.0 & 13.1/2.34/19.7 & \textbf{3.7}/0.95/\underline{60.8} & 40.7/6.72/10.6 & 6.0/1.40/44.3  \\
    & NBE+SLD \cite{do2022learning} &  \underline{6.5/0.9/38.4} & \underline{7.2/0.68/32.7} & 4.4/\underline{0.91}/\underline{53.0} & \underline{3.8}/\underline{0.94}/\textbf{66.5} & \textbf{6.0}/\underline{0.91}/\underline{40.0} & \underline{5.0/0.99/50.5} \\
    & \textbf{GSFFs-PR} Feature (34 classes) (Sgl) & \bf4.9/0.68/51 & \bf3.1/0.25/68 & \bf3.0/0.58/69 & 5.1/\underline{0.85}/49 & \underline{6.3}/\textbf{0.84}/\textbf{41} & \textbf{3.3/0.69/62} \\
    & \textcolor{gray}{\textbf{GSFFs-PR} Feature (84 classes) (Sgl)} & \textcolor{gray}{3.6/0.5/63} & \textcolor{gray}{2.7/0.2/77} & \textcolor{gray}{2.4/0.39/73} & \textcolor{gray}{3.6/0.54/63} & \textcolor{gray}{4.8/0.53/53} & \textcolor{gray}{2.5/0.42/68}  \\
    \hline
    \multirow{3}*{\begin{sideways} PPM \end{sideways}} 
    & SegLoc (Sgl) \cite{pietrantoni2023segloc} & \bd{3.9}/\bd{0.72}/51.0 &  \bd{3.2/0.37}/56.4 & \bd{4.2}/0.86/41.8 & 6.6/1.27/33.84 & \bd{5.1}/\bd{0.81}/\bd{43.1} & \bd{3.5}/\bd{0.78}/34.5 \\
   & \textbf{GSFFs-PR} Privacy (34 classes) (Sgl) & 10.2/1.8/28 &  5.8/0.49/42 & 5.8/1.2/45 & 6.5/1.2/40 & 13.9/2.32/23 & 6.5/1.5/46 \\
   & \textcolor{gray}{\textbf{GSFFs-PR} Privacy (84 Classes) (Sgl)} & \textcolor{gray}{4.6/0.67/53} & \textcolor{gray}{4.1/0.33/49} & \textcolor{gray}{4.1/0.74/55} & \textcolor{gray}{4.0/0.61/58} & \textcolor{gray}{6.6/0.92/38} & \textcolor{gray}{3.7/0.75/57} \\

    \hline
    \bottomrule
    \end{tabular}}
    \caption{Localization results on Indoor6, comparison with Non-Privacy-Preserving Methods (NPPM) and Privacy-Preserving Methods (PPM). Median position error (cm.) ($\downarrow$)/ median rotation error (°) ($\downarrow$)/ recall at 5cm/5° ($\%$) ($\uparrow)$.} 
      \label{tab:indoor6}
    \end{table*}

\PAR{Comparison with privacy-preserving approaches}
We compare the performance of the \gsfloc Privacy version against other privacy-preserving visual localization methods, including 
SegLoc \cite{pietrantoni2023segloc} and descriptor-less visual localization methods \cite{zhou2022geometry,wang2023dgc}, on
Cambridge Landmarks in 
\cref{tab:cambridge} and on  7Scenes with DSLAM pseudo GT in \cref{tab:7scenesWithDslam}.
Similar to our approach, both baseline 
try to remove the reliance on traditional high-dimensional features for visual localization and hence try to improve privacy. 
On both datasets, our pipeline proves to be much more accurate than all the other privacy-preserving methods. 
Finally, 
Tab.~\ref{tab:indoor6} shows 
results on the challenging Indoor6 dataset, which contains scenes with multiple rooms, uneven camera distributions, and large illumination changes. \gsfloc{} trained with 84 classes outperforms SegLoc in average recall, while \gsfloc{} trained with 34 classes is on par with SegLoc. 
This shows that increasing the number of classes increases the discriminative power of the segmentation, which is particularly beneficial for larger and  complex 
scenes such as \textit{scene1} and \textit{scene5}. 
The \supmat provides an 
ablation study on varying the number of classes. 

\begin{table*}[tt]
  \centering
  \begin{tabular}{l|c|c||c|c}
    & KC (F) & OH (F) & KC (P) & OH (P) \\
    \hline
    Coarse  only
    & 37.5/0.80 & 670/1.10 & 92.2/1.55 & 82.5/1.50 \\
    Fine only & 22.6/0.32 & 55.1/0.77 & 28.0/0.45 & 151/0.96 \\
    \hline
    k-means & 25.2/0.33 & 27.6/0.50 & 37.2/0.64 & 52.3/0.84 \\
    No SOpt & 23.5/0.30 & 23.4/0.45 & 31.0/0.45 & 30.8/0.57 \\
    No MVR &  20.0/0.28 &  22.5/0.49 &  29.6/0.51 &  29.1/0.49 \\
    No SE & 23.5/0.31 & 21.9/0.42 & 27.0/0.42 & 26.0/0.47 \\
    \hline
    GSFFs-PR & 17.9/0.27 & 21.4/0.41  & 24.3/0.39 & 25.6/0.49 \\
    \hline
    \bottomrule
    \end{tabular}
    \caption{Ablation of different component of our pipeline on King's College (KC) and Old Hospital (OH) for Feature (F) and Privacy (P) versions. Median position error (cm.)/  rotation error (°) ($\downarrow$).}
      \label{tab:ablations}
    
    \end{table*}

\subsection{Ablations and Discussion}
\label{sec:abl}

\cref{tab:ablations} shows results for our ablation studies, where 
we study a single component per ablation experiment.

\myparagraph{No hierarchy}
First, we evaluate the importance of the hierarchical 
approach
by either removing the fine level (Coarse Only) or by 
removing the coarse level (Fine Only). The results in 
\cref{tab:ablations} (first two rows)  
show  that alignment on the finer level yields better pose accuracy  and that increasing the convergence basin with a hierarchical approach is necessary to ensure accurate localization.

\myparagraph{Clustering} 
As shown in Tab.~\ref{tab:ablations} (k-means), when we replace the 
spectral clustering with a simple  K-means clustering to derive the prototypes in the 3D feature space, 
the accuracy drops, confirming that spectral clustering 
is a critical component of the representation learning process.

\myparagraph{Feature field learning}
The accuracy drops also when  we replace the scale-aware feature encoding with a simple point-wise projection of 3D Gaussian centers (\cf 
\cref{tab:ablations} (No SE)), or when we remove the multi-view regularization in the contrastive losses (\cf 
\cref{tab:ablations} (No MVR)). This shows that both components leverage the geometric information contained in the 3D Gaussian model, 
yielding
more robust and more discriminative representations, and hence a higher pose accuracy.

\myparagraph{Segmentation optimization}
By removing the loss from \cref{eq:ce}, we do not optimize the segmentations during training (\cref{tab:ablations} (No SOpt)).
Instead, during inference, we align soft assignments between features and prototypes. 
Due to the prototypical loss, the soft assignments are aligned well enough to provide decent accuracy, however the segmentation head provides a significant gain on accuracy.
Furthermore, the feature alignment also benefit from the segmentation optimization during training. 

\myparagraph{Sparsity of the representation}
To evaluate the role of the dense representation, 
we adopt an approach closer to SegLoc by only applying pose refinement on a sparse set of points. To do this we reproject only  Gaussian centers into the image space,  interpolate 2D extracted/3D rendered segmentations at these locations and apply \cref{eq:featalignpp} for pose refinement. In this sparse setup, the pose refinement does not converge, which suggests that a sparse signal is not sufficient when backpropagating through the renderer.

\section{Conclusion}
In this paper, we propose a Gaussian Feature Field (\gsf{}) that combines an explicit geometry representation (3DGS) and an implicit feature field into a scene representation suitable for visual localization through pose refinement. 
The resulting VL framework relies on a learned feature embedding space shared between the scale-aware \gsf{} and a 2D encoder. 
Both the field and the encoder are optimized in a self-supervised way via contrastive learning. By leveraging the geometry and structure of the GS representation, we regularize the training through multi-view consistency and clustering. The clusters are further used to  convert the features into segmentations to enable privacy-preserving localization.
Both the privacy-preserving and non-privacy-preserving variants of our localization framework outperform the relevant state-of-the-art.

{
\small
\PAR{Acknowledgments} Maxime and Torsten received funding from NAVER LABS Europe. Maxime was also supported by the the Grant Agency of the Czech Technical University in Prague (No. SGS24/057/OHK3/1T/13). We thank our colleagues for providing feedback, in particular Assia Benbihi. 
}

{
    \small
    \bibliographystyle{ieeenat_fullname}

}

\vspace{1cm}
 {\Large \bf APPENDIX} \\
 
 \appendix
 In this appendix 
first in  \cref{sec:additional}, we provide additional explanations with regard to the contrastive losses, the optimal transport association procedure, and the volumetric feature encoding. In \cref{sec:details}, we detail the training and visual localization parameters. We also provide a pseudo-algorithm describing \gsf{} training in \cref{alg:train}.
Then, in \cref{sec:morexps}, we provide visual localization results on the 12Scenes dataset with 
ablation studies and in \cref{sec:privacyattack} we present a
privacy attack experiment. 
 Finally, in \cref{sec:more_visus} we display more visualizations of rendered segmentation,  while commenting on the attached supplementary videos.

\section{Technical details of \gsf}
\label{sec:additional}

\subsection{Contrastive losses and regularization terms}

Here we provide more details regarding the derivation of the losses in \cref{sec:feature_field}.
We recall that our is training  the model in a self-supervised manner to align the feature maps 
$\bfF^\text{3D}$ and $\bfF^\text{2D}$. 
Therefore, during training at each step we sample $N$ pixels in these maps to be aligned. Let us denote the
$N$ corresponding pairs of pixel aligned extracted/rendered features
by
$\{\bfF^\text{2D}_n,\bfF^\text{3D}_n\}_{n=1}^N$. 
The contrastive loss has two terms. The first one is  a term enforcing similarity of 2D extracted features with regard to 3D rendered features,  and the second term is enforcing similarity of the 2D rendered features with regard to 3D extracted features: 
$$\textstyle{L_{NCE} = -\frac{1}{N}  \sum_{u=1}^N  \log \left( \frac{\exp{\left(\bfF^\text{3D}_{u} \cdot \bfF^\text{2D}_{u}/\tau \right)}}{\sum_{j=1}^N \exp(\bfF^\text{3D}_{u} \cdot \bfF^\text{2D}_{j}/\tau)} \right)  }$$
$$\textstyle{ -\frac{1}{N} \sum_{u=1}^N \log \left( \frac{\exp{\left(\bfF^\text{2D}_{u} \cdot \bfF^\text{3D}_{u}/\tau \right)}}{\sum_{j=1}^N \exp(\bfF^\text{2D}_{u} \cdot \bfF^\text{3D}_{j}/\tau)} \right)}$$
yielding
$L_{NCE}$ equal to:
$$\textstyle{
-\frac{1}{N}  \sum_{u=1}^N  \log \left( \frac{\exp{\left(\bfF^\text{3D}_{u} \cdot \bfF^\text{2D}_{u}/\tau \right)} \exp{\left(\bfF^\text{2D}_{u} \cdot \bfF^\text{3D}_{u}/\tau \right)}}{\sum_{j=1}^N \exp(\bfF^\text{3D}_{u} \cdot \bfF^\text{2D}_{j}/\tau)  \sum_{j=1}^N \exp(\bfF^\text{2D}_{u} \cdot \bfF^\text{3D}_{j}/\tau)} \right)} \enspace . $$
From this we can derive \cref{eq:nce},
where the normalization factor 
$A$ is:
$$\textstyle{A = \left(\sum_{j=1}^N \exp(\bfF^\text{3D}_{u} \cdot \bfF^\text{2D}_{j}/\tau) \right)  \left(\sum_{j=1}^N \exp(\bfF^\text{2D}_{u} \cdot \bfF^\text{3D}_{j}/\tau) \right)}$$

Similarly, the prototypical contrastive loss has a term enforcing  the similarity between 2D extracted features and the associated prototypes, and a second term enforcing similarity between 3D rendered features and the associated prototypes:
$$\textstyle{L_{PRO} = -\frac{1}{N} \sum_{n=1}^N \log\left(\frac{
 \exp{((\bfF^\text{3D}_{n} \cdot \bfp_{n} )/\tau )}}{\sum_{j=1}^K \exp(\bfF^\text{3D}_{n} \cdot \bfp_{j}/\tau)}\right) } $$
 $$\textstyle{
   -\frac{1}{N} \sum_{n=1}^N  \log\left(\frac{
 \exp{((\bfF^\text{2D}_{n} \cdot \bfp_{n}/\tau )}}{\sum_{j=1}^K \exp(\bfF^\text{2D}_{n} \cdot \bfp_{j}/\tau)} \right)}$$
yielding
$L_{PRO}$ equal to
$$\textstyle{-\frac{1}{N} \sum_{n=1}^N \log\left(\frac{
 \exp{((\bfF^\text{3D}_{n} \cdot \bfp_{n} )/\tau )} \exp{((\bfF^\text{2D}_{n} \bfp_{n}/\tau )}}{(\sum_{j=1}^K \exp(\bfF^\text{3D}_{n} \cdot \bfp_{j}/\tau)) (\sum_{j=1}^K \exp(\bfF^\text{2D}_{n} \cdot \bfp_{j}/\tau))}\right) }$$
which simplifies to Eq.~\textcolor{blue}{3}
with the normalization factor:
$$\textstyle{B = \left(\sum_{j=1}^K \exp(\bfF^\text{3D}_{u} \cdot \bfp_{j}/\tau) \right)  \left(\sum_{j=1}^K \exp(\bfF^\text{2D}_{u} \cdot \bfp_{j}/\tau) \right)} \enspace .$$

 \begin{algorithm*}
 \caption{Pseudo algorithm describing the training process of \gsf. }
  \label{alg:train}
      
 \SetAlgoLined
 \KwData{ $M$ posed training images, $K$ target number of classes and a total number of training iterations $N_{\textrm{iter}}$}
 Build SfM and pretrain a GoF~\cite{ZehaoTOG24GaussianOpacityFields} with the training images\\
 Apply spectral clustering on the set of 3D Gaussian centers to initialize the spatial prototypes $\{p_k\}_{k=1}^K$\\
 Then, to train the \gsf{},  iterate: \\
 \For{iter in range($N_{\textrm{iter}}$)}{
    Sample a random image $I$ and viewpoint $P$ \\
    Extract 2D image-based features $\bfF^\text{2D}$ and segmentation map $\bfS^\text{2D}$  \\
    For each 3D Gaussian $\calG_i$, extract a scale aware volumetric feature $\bfg_i$ from the triplane using the covariance kernel based encoding (cf. Sec.~\textcolor{blue}{3.1})\\
    Assign a 3D segmentation label $s_i$ to each 3D Gaussian $\calG_i$ based on volumetric feature/prototypes similarities (cf. Sec.~\textcolor{blue}{4})\\
    From the pose $P$ rasterize 3D segmentation labels and volumetric features to obtain $\bfS^\text{3D}$ and $\bfF^\text{3D}$ \\
    Sample a novel viewpoint $\widehat{P}$, find correspondences between image $I$ and $\hat{I}$
    (cf. Sec.~\textcolor{blue}{3.2}). \\
    Render 3D features/segmentation maps from $\widehat{P}$ and replace features/segmentation in $\bfF^\text{3D}$, $\bfS^\text{3D}$
    with $\widehat{\bfF}^\text{3D}$,
    $\widehat{\bfS}^\text{3D}$
    at correspondence location\\
    Repeat the replacements process for 2D features/segmentations \\
    Compute the contrastive loss $L_{NCE}$ from $\bfF^\text{2D}$ and $\bfF^\text{3D}$ \\
    Associate a prototype $p_k$ to each pair $\{\bfF^\text{2D}_i$, $\bfF^\text{3D}_i\}$ using the OT labeling procedure  (cf. \cref{sec:OTassoc}) \\
    Compute $L_{PRO}$, $L_{CE}$ from these associations as well as from $\bfF^\text{2D}$, $\bfF^\text{3D}$, $\bfS^\text{2D}$, $\bfS^\text{3D}$ (cf. Sec.~\textcolor{blue}{3.2})\\ 
    Compute regularization losses $L_{TVL}$, $L_{Depth}$ and photometric loss  $L_{PHO}$ from the rendered image \\
    Jointly update the image encoder, the triplane field and 3D Gaussians by minimizing $L = L_{PHO} + .5 L_{NCE} + .5 L_{PRO} + .5 L_{CE} + .1 L_{TVL} + 0.05 L_{Depth}$ \\
    Update the prototypes $p_k$ with an EMA scheme based on volumetric features $g_i$ and the spectral clustering assignments
    }
\end{algorithm*}

\subsection{Optimal transport (OT) associations} 
\label{sec:OTassoc}

In this section,
we detail the optimal transport association step from \cref{sec:prototypes}.
Given a batch of $N$ pairs of pixel aligned extracted/rendered features $\{\bfF^\text{2D}_n,\bfF^\text{3D}_n\}_{n=1}^N$ and a set of $K$ prototypes $P\!\in\! \Re^{K\times D}$, we aim at associating a prototype per pair of pixel aligned features. 
We want this association operation to respect two criteria: 
1) A single prototype must be associated per pair of features so that the extracted/rendered features are pushed toward the same "class" in the feature space. 2) Predictions must be as balanced as possible to avoid collapse. 

To solve these constraints, we resort to using optimal transport, where we frame this problem as finding a mapping $Q \!\in\! \Re^{N\times K}$ between pixels and prototypes that maximizes the feature similarity between the pairs of features and the prototypes. We define the joint feature/prototypes similarities  $S \in \Re^{K\times N}$ as: 
$$\textstyle{
S_{kn}  = 
\exp{\left( (\bf\text{F}^{2D}_{n} \cdot p_k + \bf\text{F}^{3D}_{n} \cdot p_k)/\tau \right) }/C} \enspace ,$$ 
with  
$$\textstyle{C=(\sum_k{\exp{(\bf\text{F}^{2D}_n \cdot p_{k}/\tau)}})(\sum_k{\exp{(\bf\text{F}^{3D}_n \cdot p_{k}/\tau)}})} \enspace .$$
We introduce the following objective 
where $Q$ maximizes the joint feature/prototypes similarities $S$:  
$$\textstyle{
    \max_{Q \in U(\frac{1}{N},\frac{1}{K})} Tr(Q (-log S)^t) + \lambda h(Q) } \enspace ,$$
where the entropy term $h(Q)$ encourages balanced predictions while using joint extracted/rendered feature prototypes similarities yields a single association per feature pair.
If we relax $Q$ such that it belongs to the transportation polytope $U(\frac{1}{N},\frac{1}{K})$ \cite{distances2013lightspeed}, $Q$ can be efficiently computed with the iterative Sinkhorn-Knopp algorithm~\cite{distances2013lightspeed}. The final associations $ \Gamma\!\in\! \Re^N$ are obtained 
with $\Gamma\!=\! \text{argmax}_k(Q)$.

\subsection{Projecting 3D Gaussians onto the triplane}

In this subsection, we explain how a volumetric feature for each 3D Gaussian is derived from the triplane grid \cref{sec:feature_field}.
The triplane grid is centered at the origin of the world coordinate space and it is composed of three orthogonal 2D planes $H_{xy},H_{xz},H_{yz} \in \Re^{D\times R \times R}$, $D$ being the triplane feature dimension and $R$ the resolution of the grid. 
We project each 3D Gaussian $\calG_i$ onto these planes and derive three Gaussian kernels 
$\calG^{xy}_i,\calG^{xz}_i,\calG^{yz}_i$ from the projections, which we use to obtain scale aware volumetric features $\bf\text{g}_i^{3D}$. 

To obtain the $xy$ feature, we proceed as follows ($xz$ and $yz$ features are obtained similarly).
Let $m_i$ by the center of $\calG_i$ and  $\Sigma_i$ its covariance matrix. We first project the center on the plane yielding $m_i^{xy}$. 
We perform orthographic projection of the covariance matrix to obtain $\Sigma_i^{xy}$.
We define a grid of dimension 5 by 5 centered on $m_i^{xy}$. On the plane $xy$, we use the coordinates of the points in the grid $u$ to define the following Gaussian kernel:
$$\textstyle{\calG^{xy}_i(\bfu) = \frac{1}{z}exp(-\frac{\bfu (\Sigma_i^{xy})^{-1} \bfu^t}{2})} \enspace ,$$
where $Z$ is a normalization constant.
We query the feature plane $H_{xy}$ for each point $u_k$ in the grid and apply the Gaussian kernel on the queried features. This yields a $D$-dimensional feature $\bf\text{g}_i^{xy}$ associated to $\calG^{xy}_i$. We repeat this operation for $H_{xz},\calG^{xz}_i$ and $H_{yz},\calG^{yz}_i$. The resulting features $\bf\text{g}_i^{xy},\bf\text{g}_i^{xz},\bf\text{g}_i^{yz}$ are summed to obtain the volumetric feature $\bf\text{g}_i^{3D}$ of $\calG_i$.

\section{Implementation details}
\label{sec:details}

\subsection{Additional training details}

Similar to GoF~\cite{ZehaoTOG24GaussianOpacityFields}, densification and pruning operations based on image space gradients are applied until iteration 15000. 
The densification interval is set to 600 iterations until iteration 7500, and reduced to 400 iterations afterward. 
To facilitate the convergence of the 3D Gaussian model, we train on images downscaled by a factor 4 until iteration 7500.
From iteration 15000, the geometric regularization losses from \cite{ZehaoTOG24GaussianOpacityFields} are  applied until the end of the training.
The triplane learning rate is set to 7e-3, while the encoder learning rate is set to 1e-4. 
The learning rates for 3D Gaussian primitives are identical to GoF~\cite{ZehaoTOG24GaussianOpacityFields}. 

\gsf{} is optimized with the Adam optimizer \cite{kingma2014adam}.
The prototypes are updated based on an exponential moving average (EMA) scheme with $\alpha=0.9995$ after each training iteration.
The temperature $\tau$ for the contrastive losses is set to 0.05.
In the \textit{Multi-view consistency} paragraph from Sec.~\textcolor{blue}{3.2}, the pixel reprojection threshold is set to 2 for the fine level, and to 4 for the coarse level.
The coarse encoder uses a Dinov2~\cite{oquab2023dinov2} pretrained backbone , followed by projection convolutional layers (convolutional layer with kernel size 1 to reduce the dimension, while maintaining the resolution) and a ConvNeXt block~\cite{liu2022convnet}. 
The fine encoder is composed of a shallow convolutional layers followed by a ConvNeXt block~\cite{liu2022convnet}. 
The segmentation heads contain convolutional layers with ReLU activations and GroupNorm~\cite{wu2018group} normalization.

We provide in \cref{alg:train} 
the pseudo algorithm describing the training process of the \gsf{}.

\subsection{Data pre-processing}

To reduce running time and the memory footprint, images are rescaled such that image width is 1024 pixels for Cambridge Landmarks and 480 pixels for Indoor6. 
The original image resolution of 640 by 480 pixels is used on 7Scenes. 
During visual localization, we use the same image resolution as the one used during training. 

Cambridge Landmarks and Indoor6 contain images with illumination changes, as such we learn an embedding per training image to capture these illumination changes during the training. These embeddings are only used during training to learn the scene representation.
On Cambridge Landmarks during training, we mask out the sky and pedestrians. 
The masks are extracted using the semantic segmentation model from \cite{RanftlICCV21VisionTransformers4DensePrediction}. As Indoor6 contain day/night images with extreme illumination changes we further apply CLAHE normalization on images. 

\subsection{Visual localization setup}
For the pose refinement, we use the Adam optimizer \cite{kingma2014adam} with coarse/fine learning rates of 0.5/0.2 on Cambridge Landmarks, 0.3/0.2 on Indoor6, and 0.2/0.1 on 7Scenes respectively. The number of refinement steps for the coarse and fine level is set to 150/300 on Cambridge Landmarks and Indoor6, and 75/150 on 7Scenes. Rendered areas with high distortion (see \cite{ZehaoTOG24GaussianOpacityFields} for the definition of distortion) are masked out during refinement. Additionally, the sky is masked out on Cambridge Landmarks.

\begin{table}
\resizebox{\linewidth}{!}{
\begin{tabular}{cc|c|c|c|c}
 
     & N Classes & KingsCollege & OldHospital & ShopFacade & StMarysChurch  \\
     \hline
      \multirow{2}*{\begin{sideways} FP \end{sideways}} 

     & 34 & 0.034 & 0.032 & 0.024 & 0.028 \\
     & 84 & 0.046 & 0.043 & 0.027 & 0.031 \\
     \hline
      \multirow{2}*{\begin{sideways} BP \end{sideways}} 

     & 34 & 0.065 & 0.065 & 0.071 & 0.076 \\
     & 84 & 0.462 & 0.401 & 0287 & 0.392 \\
     \hline
     \bottomrule
\end{tabular}}
\caption{Computation time (s) for a forward pass FP (rendering) and a backward pass BP (pose optimization) on Cambridge Landmark for \gsf{} trained with 34 classes and 84 classes.}
\label{tab:time}
\end{table}

\subsection{Training time and rendering quality}
In \cref{tab:NVS} we report novel view rendering  quality evaluated with PSNR, SSIM~\cite{CriminisiTIP04SSIM} and LPIPS~\cite{zhang2018unreasonable} image metrics as well as the corresponding  model training time for the Cambridge Landmark scenes.  
From these results in \cref{tab:NVS}, we observe that \gsf-Feature, compared to
GoF \cite{ZehaoTOG24GaussianOpacityFields}, 
 adds computational overhead (the cost of training the feature fields), but 
it does not decrease the novel view synthesis quality. 
Note that \gsf-Privacy cannot render RGB images for privacy reasons. 

We also provide  rendering time and backward pass time for the pose optimization in \cref{tab:time} for 34 and for 84 classes. We can observe that while the forward pass is only slightly increased, the cost of backward pass increases significantly with the increase the number of classes.  
Globally, through our experiments,  we found that in general using 34 classes is a good compromise between rendering speed and pose accuracy (see the accuracies in \cref{tab:cambridge} 
and 
\cref{tab:time} for the training times).

\begin{table}[tt]
    \resizebox{\linewidth}{!}{
  \begin{tabular}{l|c|c|c|c|c}
    & Training & KC & OH  & SF & SMC \\
    & time (h) & \multicolumn{4}{c}{PSNR / SSIM / LPIPS} \\
    \hline
   GoF [81] & 0.7 & 16.95/0.69/0.27 & 16.98/0.63/0.30 & 20.04/0.74/0.21 & 18.49/0.71/0.25 \\
   GSFFs & 10.8 & 16.92/0.69/0.27 & 16.94/0.61/0.30 & 20.02/0.74/0.21 & 18.47/0.71/0.26 \\
    \hline
    \bottomrule
    \end{tabular}
    }
    \caption{Evaluating novel view rendering.}
      \label{tab:NVS}
    \end{table}

\section{Additional Localization Experiments}
\label{sec:morexps}

\begin{table*}[ttt]
\scriptsize
\resizebox{\linewidth}{!}{
  \begin{tabular}{l||c|c|c|c|c|c|c|c|c|c|c|c}
     Model & kitchen & living & bed & kitchen & living & luke & gates362 & gates381 & lounge & manolis & office2 5a & office2 5b \\
    \hline
    \textbf{GSFFs-PR} Feature (34 Classes) (NV) & 0.3/0.2/99 & 0.3/0.18/100 & 0.4/0.17/100 & 0.7/0.42/91 & 0.4/0.21/96 & 0.6/0.27/97& 0.5/0.23/100 & 0.5/0.27/99 & 0.8/0.29/97 & 0.5/0.22/99 & 0.9/0.41/99 & 1.1/0.41/94 \\
   \textbf{GSFFs-PR} Privacy (34 Classes) (NV) & 0.6/0.32/97 &  0.6/0.29/100 & 0.5/0.23/100 & 0.9/0.51/75 & 0.7/0.30/99 & 0.8/0.30/96  & 0.8/0.31/98 & 0.7/0.36/98 & 1.6/0.54/91 & 0.7/0.31/97 & 1.3/0.65/95 & 1.8/0.83/69\\
   \hline
    NeRF-SCR \cite{chen2024leveraging} & 0.9/0.5 & 2.1/0.6 & \underline{1.6/0.7} & 1.2/0.5 & 2.0/\underline{0.8} & 2.6/1.0 & 2.0/0.8 & 2.7/\underline{1.2} & 1.8/\underline{0.6} & 1.6/0.7 & 2.5/0.9 & 2.6/0.8 \\
    PNeRFLoc \cite{zhao2024pnerfloc} & 1.0/0.6 & \underline{1.5/0.5} & \underline{1.2/0.5} & \textbf{0.8/0.4} & 1.4/\textbf{0.5} & 8.1/3.3 & \underline{1.6}/0.7 & 8.7/3.2 & 2.3/0.8 & 1.1/\underline{0.5} &  X & 2.8/0.9 \\
    GSPlatloc \cite{zhai2024splatloc} & \underline{0.8/0.4} & \textbf{1.1/0.4} & \underline{1.2/0.5} & \underline{1.0/0.5} & \underline{1.2}/\textbf{0.5} & 1.5/\underline{0.6} & \textbf{1.1}/\underline{0.5} & \underline{1.2}/\textbf{0.5} & \underline{1.6}/\textbf{0.5} & 1.1/\underline{0.5} & 1.4/\underline{0.6} & \textbf{1.5/0.5} \\
    NeuraLoc \cite{zhai2025neuraloc} & 0.9/0.5 & \textbf{1.1/0.4} & 1.3/0.6 & \underline{1.0}/0.6 & \underline{1.2}/\textbf{0.5} & \underline{1.4}/0.7 & \textbf{1.1}/\underline{0.5} & \textbf{1.1/0.5} & 1.7/\underline{0.6} & \underline{1.0/0.5} & \underline{1.3/0.6} & \textbf{1.5/0.5} \\
   \textbf{GSFFs-PR} Feature (34 Classes) (NV) & \textbf{0.7/0.4} & \textbf{1.1/0.4}& \textbf{1.1/0.4} & \textbf{0.8/0.4} & \textbf{1.0/0.5} & \textbf{1.3/0.5} & \textbf{1.1/0.4} & \underline{1.2}/\textbf{0.5} & \textbf{1.4/0.5} & \textbf{0.8/0.4} & \textbf{1.2/0.5} & \underline{1.7/0.6} \\
    \hline
    \bottomrule
    \end{tabular}}
    \caption{Localization results on 12Scenes (SfM pGT top rows / DSlam pGT bottom rows). Median pose error (cm.) ($\downarrow$)/ Median angle error (°) ($\downarrow$)/ Recall at 5cm/5° ($\%$) ($\uparrow)$.} 
      \label{tab:12Scenes}
    \end{table*}

\subsection{12Scenes Dataset}
\label{sec:12Scenes}

In \cref{tab:12Scenes}, we report localization results on the 12Scenes \cite{valentin2016learning} dataset for both the SfM pseudo ground truth (pGT) and the DSLAM pGT. We compare our \gsf{}-Feature model against three non privacy preserving feature rendering based-approaches methods 
NeRF-SCR \cite{chen2024leveraging},
PNeRFLoc \cite{zhao2024pnerfloc}, NeuraLoc \cite{zhai2025neuraloc} and 
GSPlatloc \cite{zhai2024splatloc}.

\gsf{} is more accurate than all baselines and clearly outperforms 
the rendering based-approaches \cite{chen2024leveraging,zhao2024pnerfloc},  while providing accuracy improvement compared to \cite{zhai2024splatloc,zhai2025neuraloc} on most of the scenes.  

\subsection{Varying the number of 
prototypes}
\label{sec:nclasses}
In this section, we study the influence of the number of classes/prototypes on both the feature- and the segmentation-based variants of our approach. 
All experiments in the main paper use a feature dimension of 16 with 34 segmentation classes. As we render both features and segmentations in the same variable (each channel is independently rendered), we  compile the rasterizer with a fixed rendering dimension of 50 (16 + 34) 
yielding a good compromise between rendering speed and discriminative power. 

\begin{figure*}[t]
\centering
\includegraphics[width=0.85\linewidth]{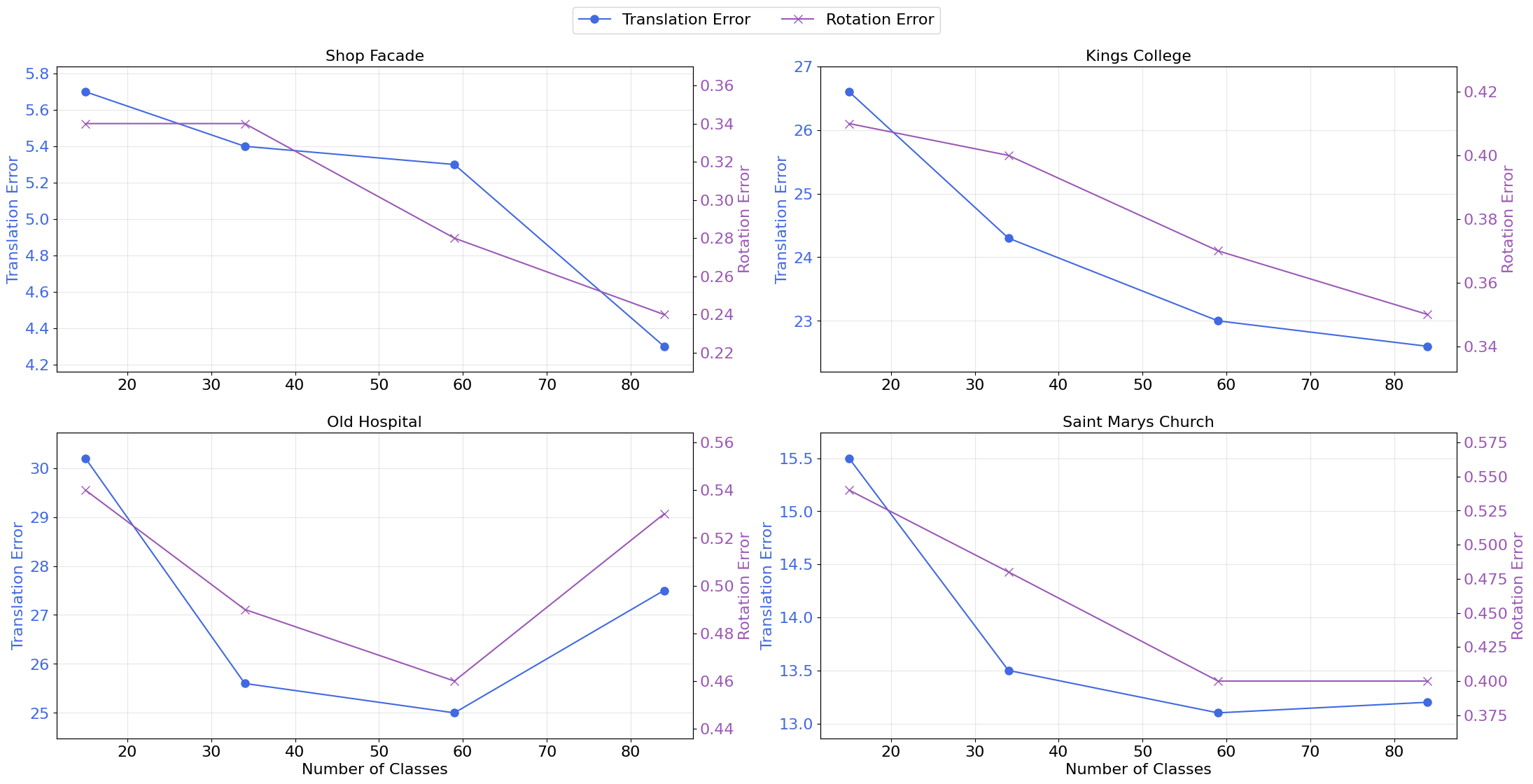}
\caption{Median pose error (cm.) ($\downarrow$) and Median angle error (°) ($\downarrow$) on Cambridge Landmarks for models trained with 15/34/59/84 classes.
}
\label{fig:abl_classes}
\end{figure*}

In this ablation, we maintain a feature dimension of 16 and vary the number of classes. We compile the rasterizer with a map dimension of 100 and then train and evaluate \gsfloc{} with 84 classes, 59 classes, and 15 classes.
In \cref{fig:abl_classes}, we plot the median translation and rotation errors on Cambridge Landmarks against the number of classes of each model. In \cref{tab:indoor6} we compare 34 classes versus 84 on the Indoor6 dataset. 
From these results,  we derive the following  observations.  Using only a few classes is not sufficient because the lack of discriminative power. 
Increasing the number of classes 
first yields a significant  gain, however above a certain limit we
observe a drop in accuracy. We suspect that the reason is over-clustering and hence more difficulty for the representation to converge during the training.  

Overall, the number of classes must be high enough to make the  segmentation discriminative enough for localization, but not too high to ensure the convergence. 
Naturally, larger scenes with diverse viewpoints will require more classes, while for simpler scenes 
it is better to  consider less classes. 
    
Furthermore, we can see from 
\cref{tab:cambridge}
that 
\gsfloc{} Feature localization pipeline also benefits from 
the increased 
number of classes as, during training, gradients are backpropagated from segmentation and feature maps and $L_{PRO}$ implicitly uses the prototypes.

\subsection{Effect of the initialization} 
In \cref{tab:different_init} we provide pose refinement results on Cambridge Landmarks with different initial poses. Note that poses estimated with ACE~\cite{brachmann2023accelerated}  and HLoc~\cite{SarlinCVPR19FromCoarsetoFineHierarchicalLocalization} are much more accurate than DenseVLAD(DV)~\cite{ToriiPAMI18247PlaceRecognitionViewSynthesis}. 
We can observe that, as expected, improving the initial pose  results in higher final \gsfloc{} localization accuracy. 
It also requires less refinement steps to converge.  
Starting from a wide baseline such as DenseVlad~~\cite{ToriiPAMI18247PlaceRecognitionViewSynthesis}
the model needs  
more refinement steps but the optimization ultimately reaches high accuracy (especially for high resolution images) showing the robustness of \gsfloc.

\begin{table*}[tt]
  \centering
  \begin{tabular}{l|c|c|c|c|c|c|c|c}
    & Init. & Image & Coarse & Runtime & KC & OH  & SF & SMC \\
    & & Res. & Fine Steps &  Query (s) & \multicolumn{4}{c}{cm / deg} \\
    \hline
    GSplatLoc \cite{sidorov2024gsplatloc} & \cite{sidorov2024gsplatloc} & ? & 350 & 3 & 27/0.46 & 20/0.71 & 5/0.36 & 16/0.61 \\
    ACE + GS-CPR \cite{liu2024gsloc} & ACE & 512 & x & 0.2 & 25/0.29 & 26/0.38 & 5/0.23 & 13/0.41 \\
    \hline
    DV & -& -&- & -& 280/5.7 & 401/7.1 & 111/7.6 & 231/8 \\
    Vanilla-GS & DV & 1024 & 150-300 & 45 & 21.9/0.34 & 22.3/0.42 & 4.5/0.27 & 11.8/0.35 \\
    GSFFs-PR & DV & 1024 & 150-300 & 45 & \bf{17.9/0.27} & \textbf{21.4}/0.41  &  \textbf{4.1}/0.26 & 10.4/0.30 \\
    GSFFs-PR & DV & 480 & 150-300 & 8.1 & 19.7/\textbf{0.27} & 21.7/\textbf{0.36}  & 4.7/\textbf{0.23}  & \bf{8.7/0.29} \\
    GSFFs-PR & DV & 480 & 50-50 & 1.8 & 74.6/1.26 & 162/2.81 & 16.4/0.73 & 90.2/2.68 \\
    \hline
     ACE & -& -& -& -& 28.0/0.4 & 31.0/0.6 & 5.0/0.3 & 18.0/0.6 \\
    GSFFs-PR & ACE & 1024 & 150-300 & 45 & \textbf{16.1/0.22} & \textbf{17.9}/0.34 & \textbf{3.9/0.18} & \textbf{7.6/0.23}  \\
    GSFFs-PR & ACE & 480 & 150-300 & 8.1 & 17.8/\textbf{0.22} & 18.7/\textbf{0.33}  & 4.7/0.21  & 8.5/0.25 \\
    GSFFs-PR & ACE & 480 & 75-150 & 4 & 18.2/0.24 & 21.1/0.35 & 4.8/0.22 & 8.5/0.25  \\
    GSFFs-PR & ACE & 480 & 50-50 & 1.8 & 19.1/0.27 & 25.4/0.49 & 5.1/0.26 & 11.1/0.35  \\
    GSFFs-PR & ACE & 480 & 25-25 & 0.9 & 19.8/0.44 & 25.6/0.66 & 5.8/0.45 & 13.2/0.53  \\
    \hline
     Hloc & -& -& -& -& 11.1/0.20 & 15.5/0.31 & 4.4/0.2 & 7.1/0.24 \\
    GSFFs-PR & Hloc & 1024 & 0-50 & 5 & \textbf{10.9/0.18} & \textbf{14.1/0.29} & \textbf{3.9}/0.19 & \textbf{6.4/0.21} \\
    GSFFs-PR & Hloc & 480 & 0-50 & 0.9 & 11.0/0.19 & 14.2/0.30 & 3.9/\textbf{0.18} & \textbf{6.4}/0.22 \\
    \hline
    
    \bottomrule
    \end{tabular}
    \caption{Varying initialization and image resolution.  }
    \vspace{0.5cm}
   \label{tab:different_init}
    \end{table*}

\begin{figure*}[ttt]
\centering
\includegraphics[width=0.95\linewidth]{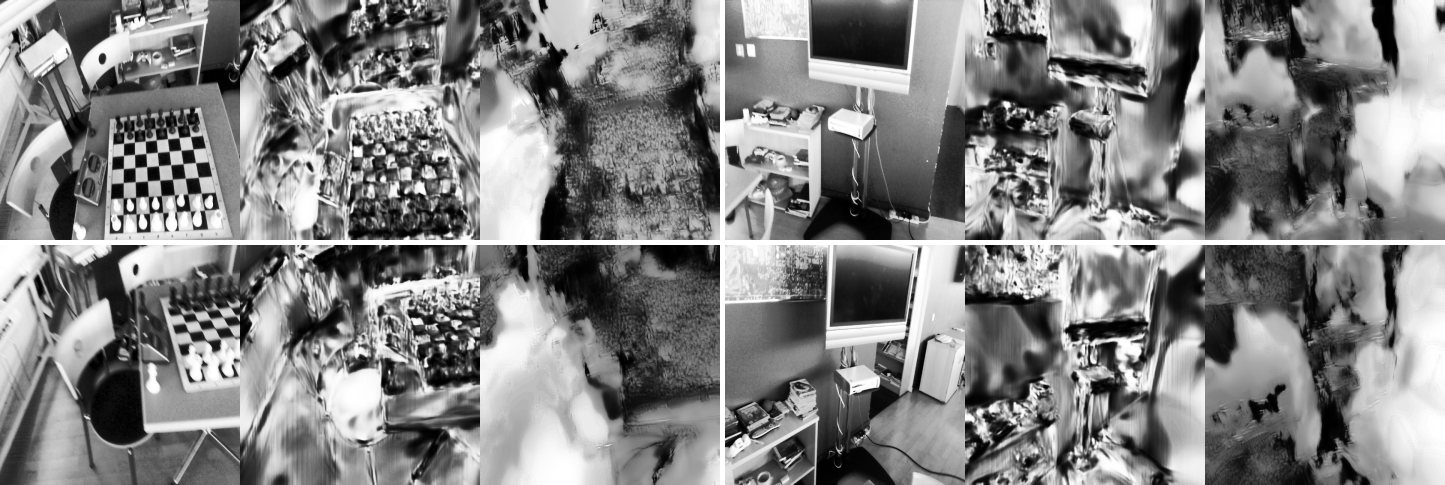}
\caption{
Left to right: original image, image inversion attack from rendering our features (middle, GSFFs-PR Feature) or rendering our segmentation (right, GSFFs-PR Privacy). }
\label{fig:privacy}
\end{figure*}

\subsection{Varying resolution and refinement steps}

In \cref{tab:different_init} we also provide results with different  coarse/fine number of refinement steps, image resolution and runtime per query.
We can see that performances 
on high resolution images
is slightly better than on low resolution images,  but this comes at a higher inference cost.  We can further decrease the running time by decreasing the refinement steps. The loss in performance is relatively small conditioned that  we
start from a good initialization.  
This suggest that we could further increase the localization speed 
by combining  low and hight resolution based refinement and stopping the optimization earlier.

\section{Privacy attack}
\label{sec:privacyattack}

To visualize and assess the degree of privacy of \gsfloc{}-Privacy, we train an inversion model \cite{pittaluga2019revealing,pietrantoni2023segloc} to reconstruct images from rendered segmentations. As a baseline, we train another inversion model to reconstruct images from feature maps rendered from \gsfloc{}-Feature. Both inversions model are trained on 6 scenes (\textit{fire}, \textit{heads}, \textit{office}, \textit{pumpkin}, \textit{redkitchen}, \textit{stairs}) from the 7Scenes dataset and evaluated the remaining scene \textit{chess}. Example reconstructed images from both \gsfloc{}-Feature and \gsfloc{}-Privacy and displayed in~\cref{fig:privacy}. We can observe that images reconstructed from \gsfloc{}-Feature reveal a lot of scenes details,  while images reconstructed from \gsfloc{}-Privacy totally obfuscate privacy sensitive information.

\section{Visualizations}
\label{sec:more_visus}
\begin{figure*}[t]
\centering
\includegraphics[width=0.9\linewidth]{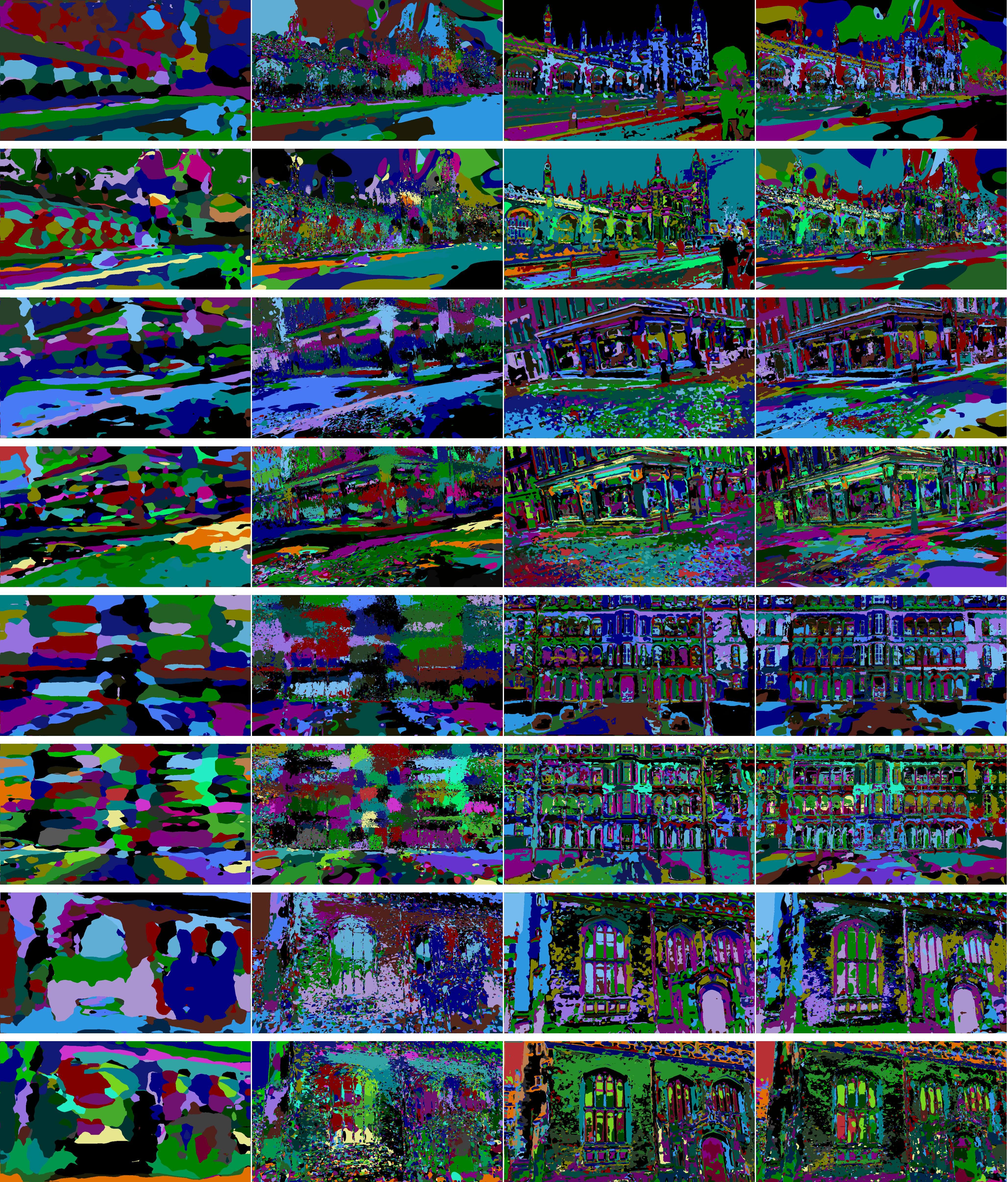}
\caption{From left to right, coarse encoder/rendered segmentation, fine encoder/rendered segmentation. Comparison between models trained with 34 classes (line 1/3/5/7) and models trained with 84 classes (line 2/4/6/8). 2D extracted and 3D rendered segmentations are well aligned which allows for accurate privacy preserving visual localization.
}
\label{fig:seg3484}
\end{figure*}

We show in \cref{fig:seg3484} pairs of 2D extracted / 3D rendered segmentations for the coarse and fine levels while comparing segmentations from models trained with 34 and 84 classes. The segmentation classes of the coarse level capture larger and less sharply defined segments in the image compared to the fine level segmentations. As shown in the ablation in \cref{tab:ablations},
this allows us to increase the convergence basin of the pose refinement, while improving fine accuracy. Using 84 classes instead of 34 results in visually finer-grained segmentations  in turn allows for more accurate pose refinement.

\end{document}